\documentclass[11pt]{article}

\usepackage[letterpaper,margin=1in]{geometry}
\usepackage[T1]{fontenc}
\usepackage[utf8]{inputenc}
\usepackage{lmodern}
\usepackage{microtype}
\usepackage{graphicx}
\usepackage{tabularx}
\usepackage{parskip}
\usepackage[numbers,square,sort&compress]{natbib}

\usepackage{amsmath,amsfonts,bm}

\def\eqref#1{equation~\ref{#1}}

\def\1{\bm{1}}

\DeclareMathAlphabet{\mathsfit}{\encodingdefault}{\sfdefault}{m}{sl}
\SetMathAlphabet{\mathsfit}{bold}{\encodingdefault}{\sfdefault}{bx}{n}

\newcommand{\softmax}{\mathrm{softmax}}

\usepackage{hyperref}
\hypersetup{colorlinks=true,linkcolor=blue!60!black,citecolor=blue!60!black,urlcolor=blue!60!black}
\usepackage{url}
\usepackage{booktabs}
\usepackage{float}
\usepackage{subcaption}
\usepackage{tikz}
\usetikzlibrary{arrows.meta,positioning,fit,backgrounds,calc,decorations.pathreplacing}
\usepackage{xcolor}
\definecolor{cstroke}{HTML}{3F4A5A}
\definecolor{cnorm}{HTML}{EDF1F6}
\definecolor{cattn}{HTML}{D9E4F2}
\definecolor{cmlp}{HTML}{DCEBDD}
\definecolor{cemb}{HTML}{F6EBDD}
\definecolor{cres}{HTML}{8A94A6}
\definecolor{csym}{HTML}{7A4B8A}
\usepackage{amsmath}
\usepackage{mathtools}
\usepackage{amssymb}
\usepackage{amsthm}
\usepackage[capitalise,nameinlink]{cleveref}
\newtheorem{theorem}{Theorem}
\newtheorem{lemma}{Lemma}
\theoremstyle{definition}
\newtheorem{definition}{Definition}
\theoremstyle{remark}

\theoremstyle{plain}

\usepackage{amsmath}
\usepackage{scalerel}
\DeclareMathOperator*{\bigtimes}{\scalerel*{\times}{\sum}}
\usepackage{tikz}
\usetikzlibrary{arrows.meta}
\DeclareRobustCommand{\crossarrow}{\mathrel{\vcenter{\hbox{\tikz[line width=0.3pt, >={To[scale=0.55]}]{
  \draw[->] (0,0)      -- (0.67em,0.4em);
  \draw[->] (0,0.4em) -- (0.67em,0);}}}}}
\DeclareRobustCommand{\pararrow}{\mathrel{\vcenter{\hbox{\scalebox{0.75}{$\rightrightarrows$}}}}}

\title{Graph Metanetworks for Transformer Based Language Models}

\author{%
  \normalsize
  \textbf{Yonathan Wolanowsky}$^{1}$ \quad \textbf{Guy Bar-Shalom}$^{1}$ \quad
  \textbf{Fabrizio Frasca}$^{1}$ \quad \textbf{Haggai Maron}$^{1,2}$\\[0.3em]
  \small
  $^{1}$Technion -- Israel Institute of Technology \qquad $^{2}$NVIDIA Research
}
\date{}

\begin{document}

\maketitle

\begin{abstract}
\noindent\textit{Weight-space learning} treats the weights of neural networks as data, processed by
other models called \textit{metanetworks}. A key challenge is that different weights can
realize the same function, related by a group of symmetries the metanetwork should respect.
While symmetry-aware metanetworks for simple architectures such as MLPs are well developed, a
comprehensive treatment of transformers, and in particular transformer-based
language models, is still missing. Importantly, their weight space has a richer symmetry group:
beyond neuron permutations, attention introduces continuous general-linear symmetries of the
attention matrices and permutations of entire heads. Existing methods for transformer weights
either capture only part of these symmetries or are restricted to a single, fixed architecture,
and none accounts for the symmetries of language-model-specific components such as the token embeddings. Building
on graph metanetworks, we propose a graph-based approach that is both symmetry-aware and
applicable across depths and widths. We formalize a weight-space symmetry group of complete
decoder-only language models and provide a graph construction that is invariant to it. We
further introduce two new transformer-population datasets with novel tasks, such as
training-corpus prediction and fine-tuning detection, on which our approach notably
outperforms existing baselines.
\end{abstract}

\section{Introduction}
The success of deep learning, together with the large number of publicly available trained
models, has created a new data modality: the weights of neural networks themselves. 
The emerging field of \textit{weight-space learning} \citep{han2026weightspace,wang2026position} treats these weights
as data to train \textit{metanetworks} that can process other networks' weights and predict model properties or edited weights with a single forward pass~\citep{unterthiner2020predicting,navon2023equivariant,liu2025metapruning}. 

Two important challenges arise when designing effective metanetworks. First, input architectures may vary widely: an MLP, a CNN, and
a transformer share no common parameterization, and even networks of the same family differ in architectural choices such as depth and width. Second, many different weight configurations realize the same function,
related by a group of symmetries acting on the weight space, most prominently the permutations
of hidden neurons \citep{navon2023equivariant,zhou2023permutation}. A metanetwork that is not
invariant or equivariant to this group would need to use its capacity and data only to learn to treat such
configurations as equivalent. These obstacles have motivated recent work on 
\textit{Graph metanetworks} \citep{lim2024graph,kofinas2024graph}, which address both challenges by
representing a neural network as a graph (with neurons as nodes and parameters as edges) and by processing
it with a graph neural network (GNN). Crucially, since GNNs
are equivariant to node permutations, they respect the neuron-permutation symmetries of the
encoded network.

Despite this progress, transformers \citep{vaswani2017attention} and, in particular, the
decoder-only language models built from them, remain underexplored as a weight space, especially in terms of the underlying symmetry structure and graph representation. These architectures arguably represent the most important and widely used today, and their weight
space is, in fact, also considerably richer in symmetries than that of other architectures. For example, within each attention head, the query and key
projections enter the computation only through their product $\mathbf{W}^{q}(\mathbf{W}^{k})^{\top}$,
and the value and output projections only through $\mathbf{W}^{v}\mathbf{W}^{O}$, so each pair
admits a continuous general-linear symmetry, on top of which entire heads can be permuted.
Existing approaches to transformer weights follow two lines, each capturing only part of this
picture. The first Graph Metanetworks \citep{lim2024graph,kofinas2024graph} provide a preliminary, simplified
construction for attention, but do not account for the general-linear symmetries and do not encode the head structure faithfully. \citet{knyazev2025accelerating}, aiming at accelerating training, refine this construction to respect head permutations, but remain restricted to permutation symmetries. Conversely, \citet{tran2025transformer} design layers equivariant to
the general-linear and head-permutation symmetries, but tie the metanetwork to a single, fixed
transformer architecture. Moreover, none of these works treats the symmetries induced by the components specific
to language models, such as the vocabulary permutation coupling the token embedding and unembedding matrices. As a result, no existing approach respects the symmetries of a complete transformer
language model while remaining agnostic to its depth, width, and number of heads.

\textbf{Our approach.} We begin by formalizing a weight-space symmetry group of a
complete decoder-only transformer language model, from the token embedding, through the stack
of transformer blocks with their normalization layers and residual connections, up to the
unembedding layer. This group combines permutations of the feed-forward hidden neurons, a permutation
of the residual-stream features shared across all blocks, per-head general-linear symmetries of
the query--key and value--output pairs, permutations of the heads within each block, as well as a
permutation of the vocabulary that couples the embedding and unembedding matrices together.
We prove that these transformations leave the function computed by the model unchanged (up to
a relabeling of the vocabulary). Our approach thus unifies a range of previous, independent subcomponent analyses \cite{tran2024monomial},\cite{kofinas2024graph},\cite{lim2024graph},\cite{knyazev2025accelerating} into a single, comprehensive framework, extending the attention symmetries by
\citet{tran2025transformer} to a complete language model, including its residual stream,
normalization layers, embedding, and unembedding.

We then design tailored graph constructions for transformers whose symmetries match this group, which we
call \emph{Graph Metanetworks for Transformers} (GMT). As in prior graph metanetworks,
parameters are placed on edges between layers of nodes, so that a single GNN can process
transformers of any depth, width, and number of heads. GMT uses
parameter-free \emph{coupling edges} and auxiliary \emph{head nodes} which tie together exactly those node
permutations that correspond to symmetries of the network: the residual-stream permutation is
shared across blocks, the dimensions of each head module are bound together within each head, and the embedding is
linked to the unembedding. We prove that, under mild rank conditions, two weight configurations
yield isomorphic graphs if and only if they are related by an element of the aforementioned group: processing these graphs with a GNN thus guarantees the metanetwork remains invariant (or equivariant), to the transformer weight space group by construction. We propose two
variants that trade exactness for efficiency: GMT-GL captures the full group, including the
general-linear symmetries, and GMT-P, which as we will show, restricts these symmetries to permutations, yielding a more compact graph.

Finally, as open-weight language models are increasingly released and modified by third
parties, questions such as \emph{what data was a model trained on?} and \emph{has it been
altered, and where?} become pressing, yet lie beyond existing weight-space benchmarks, which
focus on implicit neural representations and small vision classifiers. As a first step, we
introduce two datasets of transformer language models with tasks that serve as small-scale,
controlled proxies for these questions: predicting a model's training-corpus mixture, detecting
fine-tuning, and identifying which weights it modified. GMT outperforms existing baselines on
all three, e.g., reducing the KL divergence of corpus-mixture prediction by 25\% and halving
the error of weight-level fine-tuning detection relative to prior graph metanetworks.

\textbf{Contributions.} We (i) formalize a weight-space symmetry group of complete decoder-only
language models; (ii) introduce GMT, architecture-agnostic graph constructions whose
isomorphisms match this group; and (iii) release two transformer-population datasets with
auditing-motivated tasks, on which GMT outperforms existing baselines.
\section{Related work}
In weight-space learning, equivariant metanetworks were developed for the neuron permutations of MLPs and CNNs \citep{navon2023equivariant,zhou2023permutation} and extended to further symmetries and architectures \citep{zhou2024universal,kalogeropoulos2024scalegmn}, with graph metanetworks \citep{lim2024graph,kofinas2024graph} encoding networks as graphs processed by GNNs. For transformers, the general-linear symmetries of attention have been studied for linear mode
connectivity \citep{theus2025generalized} and, in analogous form, for processing low-rank
adapters \citep{putterman2024learning}. \citet{tran2025transformer} design layers equivariant
to the attention symmetries, but for a single, fixed architecture. Existing graph metanetworks
encode attention layers without the general-linear symmetries:
\citet{kofinas2024graph} share nodes across the
query, key, and value projections of all heads, tying the query--key and value--output
permutations and admitting spurious exchanges of dimensions across heads.
\citet{lim2024graph} do the same, and suggest, but do not use, head-index features, which in turn
violates the head-permutation symmetry in the constructed graph.
\citet{knyazev2025accelerating} partially address these issues by adding head nodes and separating the query--key from the
value--output dimensions, so that on the transformer blocks their graph captures the same
permutation symmetries as our GMT-P; they do not, however, consider the general-linear symmetries.
None of these works
addresses the symmetries of a complete language model, including the vocabulary permutation
coupling its embedding and unembedding.
We provide an extended discussion in \Cref{app:ext_prev}.

\section{Method}

In this section, we formulate our approach. We first characterize the group of weight-space symmetries of
a decoder-only transformer language model (\Cref{sec:method-symmetries}).
We then design a graph construction whose structure encodes this group,
so that a graph metanetwork applied to it is
invariant, or equivariant, to it by construction and describe the metanetwork we apply on it (\Cref{sec:method-graph}).

\subsection{Transformer Weight-Space and Symmetry Group}
\label{sec:method-symmetries}


\textbf{Transformer LM Computation and Weight Space.}
A transformer language model starts by mapping tokens in a length-$N$ input sequence to vectors $\mathbf{X}^{0} \in \mathbb{R}^{N \times d}$. This is performed by an embedding matrix $\mathbf{E} \in \mathcal{E} = \mathbb{R}^{V \times d}$, which we define to be the embedding weight space.
The initial representation $\mathbf{X}^{0}$ is then passed through $L$ transformer blocks stacked together, 
$\mathbf{X}^{\ell} = \mathrm{Block}_{\ell}(\mathbf{X}^{\ell-1})$ for $\ell = \{1,
\dots, L\}$. Each transformer block is composed of two sub-layers, a \textit{multi-head
attention (MHA)} mechanism followed by a \textit{feed-forward (FF)} network, each wrapped in a
normalization layer and a residual connection.
A normalization layer is parameterized by a scale and a shift $\boldsymbol{\gamma},
\boldsymbol{\beta} \in \mathbb{R}^{d}$; we define the normalization weight space as
$\mathcal{N} = \mathbb{R}^{d} \times \mathbb{R}^{d}$, so block $\ell$ holds a
pre-attention pair $\boldsymbol{\gamma}^{a,\ell}, \boldsymbol{\beta}^{a,\ell} \in \mathcal{N}$ and
a pre-MLP pair $\boldsymbol{\gamma}^{b,\ell}, \boldsymbol{\beta}^{b,\ell} \in \mathcal{N}$. The MHA module
\citep{vaswani2017attention} is defined as:
\begin{equation*}
\mathrm{MHA}(\mathbf{X}) = \Biggl( \bigoplus_{h=1}^{H} \xi \!\left(
\frac{\mathbf{X}\mathbf{W}^{q,h}(\mathbf{W}^{k,h})^{\top}\mathbf{X}^{\top}}{\sqrt{d_{k}}} \right) \mathbf{X}\mathbf{W}^{v,h} \Biggr) \mathbf{W}^{O},
\end{equation*}
where $\xi \equiv \text{softmax}$, $\bigoplus$ denotes concatenation across the $H$ heads, $h$ denotes head index, and the
per-head query, key and value projections $\mathbf{W}^{q,h}, \mathbf{W}^{k,h} \in \mathbb{R}^{d
\times d_{k}}$, $\mathbf{W}^{v,h} \in \mathbb{R}^{d \times d_{v}}$ together with the output
projection $\mathbf{W}^{O} \in \mathbb{R}^{Hd_{v} \times d}$ are its trainable parameters. We
define the attention weight space to be $\mathcal{A} = \big(\mathbb{R}^{d \times d_{k}}\big)^{H}
\times \big(\mathbb{R}^{d \times d_{k}}\big)^{H} \times \big(\mathbb{R}^{d \times
d_{v}}\big)^{H} \times \mathbb{R}^{Hd_{v} \times d}$.
The FF sub-layer then processes each token's
representation, typically through a two-layer MLP with a pointwise nonlinearity,
$\mathbf{W}^{1,\ell} \in \mathbb{R}^{d \times d_{\mathrm{ff}}}$ and $\mathbf{W}^{2,\ell} \in
\mathbb{R}^{d_{\mathrm{ff}} \times d}$; we define the FF weight space to be $\mathcal{F} =
\mathbb{R}^{d \times d_{\mathrm{ff}}} \times \mathbb{R}^{d_{\mathrm{ff}} \times d}$. Finally, after
the $L$ blocks, a final normalization $\boldsymbol{\gamma}^{f}, \boldsymbol{\beta}^{f} \in
\mathcal{N}$ is applied, and an unembedding linear layer $\mathbf{F} \in \mathbb{R}^{d \times V}$
maps $\mathbf{X}^{L}$ to a distribution over the vocabulary for the next token; we define the
unembedding weight space as $\mathcal{U} = \mathbb{R}^{d \times V}$.
Collecting these spaces gives the weight
space of all transformers with $L$ blocks and $H$ heads:
\begin{equation}
\label{eq:weight-space}
\mathcal{W} \!=\!\! \underbrace{\mathcal{E}}_{\text{Embed.}} \! \bigtimes_{\ell=1}^{L}
\underbrace{\Big( \mathcal{N} \,\times\, \mathcal{A} \,\times\, \mathcal{N} \,\times\, \mathcal{F}
\Big)}_{\text{Transformer Block}} \\
\times\, \underbrace{\mathcal{N} \,\times\,
\mathcal{U}}_{\substack{\text{Norm +}\\\text{Unemb.}}}
\end{equation}
where each point $w \in \mathcal{W}$ determines a function from token sequences
to next-token distributions.

\textbf{Transformer weight space symmetries.}
Two transformers with different weights may compute the same function, i.e., the same
function admits many equivalent parameterizations. In most weight-space tasks, the target does
not depend on which of these parameterizations is given, so a weight-space model should be
invariant or equivariant to the weight transformations relating them. We formalize such
transformations as a group acting on $\mathcal{W}$ (\Cref{eq:weight-space}), examining each
component of the network in turn; \Cref{tab:symmetries} summarizes the result.

(1) \textit{The embedding matrix} $\mathbf{E}$ admits a row permutation $P_{V} \in S_{V}$ --- corresponding to
permuting the vocabulary dictionary --- and a column permutation $P_{d} \in S_{d}$ --- corresponding to reindexing the features of the embedded vectors. That is, $\mathbf{E} \mapsto P_{V}^{\top} \mathbf{E} P_{d}$.
As we will show, all upcoming layers are equivariant under the action of $P_d$,
allowing us to perform this group action while carrying it through the network.

(2) \textit{The pre-attention normalization} maps each token representation $\mathbf{x} \in \mathbb{R}^{d}$ (a row of $\mathbf{X}^{\ell-1}$) to $\mathrm{Norm}(\mathbf{x}; \boldsymbol{\gamma}^{a,\ell}, \boldsymbol{\beta}^{a,\ell}) =
\frac{\mathbf{x} - \mu(\mathbf{x})\mathbf{1}^{\top}}{\sigma(\mathbf{x})} \odot \boldsymbol{\gamma}^{a,\ell} + \boldsymbol{\beta}^{a,\ell}$,
where $\mathbf{1}$ is the all-ones 
vector and $\odot$ is the entrywise product.
Note that $\mu(\mathbf{x}),\sigma(\mathbf{x})$ are permutation invariant. 
In order for the norm to respect the column permutation $P_d$, we need to transform the scale and shift parameters accordingly, i.e.\
$\boldsymbol{\gamma}^{a,\ell} \mapsto \boldsymbol{\gamma}^{a,\ell}P_{d}$, $\boldsymbol{\beta}^{a,\ell} \mapsto \boldsymbol{\beta}^{a,\ell}P_{d}$.
This gives
$\mathrm{Norm}(\mathbf{x}P_{d};\, \boldsymbol{\gamma}^{a,\ell}P_{d},\, \boldsymbol{\beta}^{a,\ell}P_{d})
= \mathrm{Norm}(\mathbf{x}; \boldsymbol{\gamma}^{a,\ell}, \boldsymbol{\beta}^{a,\ell})\, P_{d}$. \

(3) \textit{Each attention head's query and key projections} enter
the network only through the product $\mathbf{W}^{q,h}(\mathbf{W}^{k,h})^{\top}$; substituting any
invertible $G_{qk}^{h} \in \mathrm{GL}(d_{k})$ as $\mathbf{W}^{q,h} \mapsto \mathbf{W}^{q,h}
G_{qk}^{h}$, $\mathbf{W}^{k,h} \mapsto \mathbf{W}^{k,h} (G_{qk}^{h})^{-\top}$ leaves this product,
and hence the attention calculation, unchanged.
Note that the output projection $\mathbf{W}^{O} \in \mathbb{R}^{Hd_{v} \times d}$ is likewise a stack of $H$
blocks $\mathbf{W}^{O}_{h} \in \mathbb{R}^{d_{v} \times d}$, so concatenating the
per-head values and multiplying by $\mathbf{W}^{O}$ can be expressed as the following sum:
\begin{equation}
\label{eq:mha-as-a-sum}
\sum_{h=1}^{H} \xi\!\left(
\frac{\mathbf{X}\mathbf{W}^{q,h}(\mathbf{W}^{k,h})^{\top}\mathbf{X}^{\top}}{\sqrt{d_{k}}} \right)
\mathbf{X}\mathbf{W}^{v,h}\mathbf{W}^{O}_{h}.\!\!\!\!
\end{equation}
The value/output pair thus enter the network only through their
product, imposing a second general-linear symmetry $G_{vo}^{h} \in \mathrm{GL}(d_{v})$. \Cref{eq:mha-as-a-sum} only ever sums the $H$ heads, so they may in addition be
permuted by any $\sigma_{\ell} \in S_{H}$. Together, these give
the \emph{wreath product} \cite{wang2020equivariant} $\bigl(\mathrm{GL}(d_{k}) \times \mathrm{GL}(d_{v})\bigr) \wr S_{H}$.
Also
the \textit{MHA} input still carries $P_{d}$, enforcing the attention weights matrices to also be transformed by
$P_{d}$ on the left. 

(4) \textit{The pre-MLP normalization} applies the same $\mathrm{Norm}$ defined above to a seperate
scale and shift, so again it is equivariant to $\boldsymbol{\gamma}^{b,\ell}, \boldsymbol{\beta}^{b,\ell} \mapsto
\boldsymbol{\gamma}^{b,\ell}P_{d}, \boldsymbol{\beta}^{b,\ell}P_{d}$.

(5) \textit{In the feed-forward sub-layer}, the pointwise nonlinearity is applied element-wise,
so it commutes with permuting the columns of $\mathbf{W}^{1,\ell}$ and undoing
it with $(\hat{P}^{\ell})^{\top}$ on the rows of $\mathbf{W}^{2,\ell}$.
Also here the input carries $P_{d}$ (from the embedding column permutation),
enforcing us to apply $P_{d}$ on each of the \textit{ff} weight matrices.

(6) \textit{The final normalization}, applied after the last block, is governed by the $\mathrm{Norm}$ computation: the same considerations as in (2) shows the same equivariance.

(7) \textit{For the unembedding matrix} $\mathbf{F}$, 
we want to undo the permutation $P_{d}$ that we carry along the network. Additionally, since the embedding was permuted by $P_{V}$, the unembedding must also be permuted by $P_{V}^{\top}$
to match the token relabeling. The transformation to $\mathbf{F}$ is hence defined as $\mathbf{F} \mapsto P_{d}^{\top} \mathbf{F} P_{V}$.

\begin{table}[t]
\caption{The symmetry group action over a weight space $w \in \mathcal{W}$.
The attention group is
$S_{\wr} \coloneqq S_{d} \times \bigl[\bigl(\mathrm{GL}(d_{k}) \times \mathrm{GL}(d_{v})\bigr) \wr S_{H}\bigr]$.
The action column omits the layer index $\ell$, which is left unchanged.}
\label{tab:symmetries}
\begin{center}
\small
\setlength{\tabcolsep}{10pt}
\begin{tabular}{@{}llll@{}}
\toprule
\bf Network Block & \bf Weight Matrix & \bf Group Action & \bf Group Element \\
\midrule
(1) Embed. & $\mathbf{E}$ & $P_{V}^{\top}\, \mathbf{E}\, P_{d}$
  & $S_{V} \times S_{d}$ \\
\addlinespace[2pt]
\midrule
\addlinespace[2pt]
(2) Pre-attn. & $\boldsymbol{\gamma}^{a,\ell}$ & $\boldsymbol{\gamma}^{a} P_{d}$
  & $S_{d}$ \\
\phantom{(2)} norm & $\boldsymbol{\beta}^{a,\ell}$ & $\boldsymbol{\beta}^{a} P_{d}$ & \\
\addlinespace[2pt]
\midrule
\addlinespace[2pt]
(3) Attention & $\mathbf{W}^{q,h,\ell}$
  & $P_{d}^{\top}\, \mathbf{W}^{q,\sigma(h)}\, G_{qk}^{\sigma(h)}$
  & $S_{\wr}$ \\
 & $\mathbf{W}^{k,h,\ell}$
  & $P_{d}^{\top}\, \mathbf{W}^{k,\sigma(h)}\, (G_{qk}^{\sigma(h)})^{-\top}$ & \\
 & $\mathbf{W}^{v,h,\ell}$
  & $P_{d}^{\top}\, \mathbf{W}^{v,\sigma(h)}\, G_{vo}^{\sigma(h)}$ & \\
 & $\mathbf{W}^{O,\ell}_{h}$
  & $(G_{vo}^{\sigma(h)})^{-1}\, \mathbf{W}^{O}_{\sigma(h)}\, P_{d}$ & \\
\addlinespace[2pt]
\midrule
\addlinespace[2pt]
(4) Pre-MLP & $\boldsymbol{\gamma}^{b,\ell}$ & $\boldsymbol{\gamma}^{b} P_{d}$
  & $S_{d}$ \\
\phantom{(4)} norm & $\boldsymbol{\beta}^{b,\ell}$ & $\boldsymbol{\beta}^{b} P_{d}$ & \\
\addlinespace[2pt]
\midrule
\addlinespace[2pt]
(5) MLP & $\mathbf{W}^{1,\ell}$ & $P_{d}^{\top} \mathbf{W}^{1} \hat{P}$
  & $S_{d} \times S_{d_{\mathrm{ff}}}$ \\
 & $\mathbf{W}^{2,\ell}$ & $\hat{P}^{\top} \mathbf{W}^{2} P_{d}$ & \\
\addlinespace[2pt]
\midrule
\addlinespace[2pt]
(6) Final & $\boldsymbol{\gamma}^{f}$ & $\boldsymbol{\gamma}^{f} P_{d}$ & $S_{d}$ \\
\phantom{(6)} norm & $\boldsymbol{\beta}^{f}$ & $\boldsymbol{\beta}^{f} P_{d}$ & \\
\addlinespace[2pt]
\midrule
\addlinespace[2pt]
(7) Unembed. & $\mathbf{F}$ & $P_{d}^{\top} \mathbf{F}\, P_{V}$ & $S_{d} \times S_{V}$ \\
\bottomrule
\end{tabular}
\end{center}
\end{table}

\textbf{Full symmetry group.} The per-block transformations are summarized in \Cref{tab:symmetries}. Combining them gives the symmetry group of the full weight space:
\begin{equation}\label{eq:full-group}
  \mathcal{G} \!=\! \underbrace{S_{V} \times S_{d}}_{\text{Embedding}} \bigtimes_{\ell=1}^{L}
  \Bigl[\, \underbrace{\bigl(\mathrm{GL}(d_{k}) \times \mathrm{GL}(d_{v})\bigr) \wr S_{H}}
            _{\text{attention}}
  \;\times\; \underbrace{S_{d_{\mathrm{ff}}}}_{\text{FF}} \Bigr].
\end{equation}
We summarize the discussed transformer symmetries with the following theorem, proved in~\Cref{app:symmetries-proof}.
\begin{theorem} [the elements of $\mathcal{G}$ are symmetries of the weight space]
\label{thm:symmetries}
Let $f_{w}$ denote the function computed by a decoder-only transformer language model with weights
$w \in \mathcal{W}$.
Then $f_w$ is invariant to the group $\mathcal{G}$ with the action defined in \Cref{tab:symmetries}.
i.e, for $w \in \mathcal{W}, g \in \mathcal{G}$, it holds that $f_{g \cdot w} = f_{w}$.
\end{theorem}

\textbf{Additional symmetries.}
$\mathcal{G}$ is not the maximal symmetry group of a transformer language model. For example,
each normalization layer admits a diagonal rescaling
$\boldsymbol{\gamma}, \boldsymbol{\beta}, \mathbf{W} \mapsto \boldsymbol{\gamma}D, \boldsymbol{\beta}D, D^{-1}\mathbf{W}$ where $\mathbf{W}$ is a linear map followed by it.
In addition, some activation functions induce scaling or sign symmetries of the feed-forward
neurons \citep{godfrey2022symmetries,tran2024monomial,kalogeropoulos2024scalegmn}. These
symmetries are complementary to the ones we treat and can be handled on top of our
constructions, e.g., by canonicalizing the weights; we leave this to future work.

\subsection{Graph Construction}
\label{sec:method-graph}
Having identified the symmetry group $\mathcal{G}$ (\Cref{eq:full-group}), we seek a model
that is invariant or equivariant to it. Architectures tailored to $\mathcal{G}$ typically tie
their input dimension to a single transformer architecture, and thus cannot be applied across
depths $L$, widths $d$, or numbers of heads $H$. Following \citet{lim2024graph} and
\citet{kofinas2024graph}, we instead represent a transformer as a \emph{graph} whose edges
carry the model's parameters, augmented with structural ``gadgets'' that encode its
symmetries, and process it with a graph neural network (GNN). Since GNNs apply to graphs of
any size and are equivariant to node permutations, invariance or equivariance to $\mathcal{G}$
follows from the graph construction alone. Unlike ours, the earlier constructions of \cite{lim2024graph,kofinas2024graph} do not
match $\mathcal{G}$: they do not model the general-linear symmetries and do not faithfully encode the head structure.
The construction of \citet{knyazev2025accelerating} models the head permutations $S_{H}$, but restricts each per-head $\mathrm{GL}(d_{k}) \times \mathrm{GL}(d_{v})$ to its permutation subgroup $S_{d_{k}} \times S_{d_{v}}$, and does not encode the vocabulary permutation coupling the embedding and unembedding.

\textbf{GL-Invariant Graph Construction (GMT-GL).}
\label{sec:GL-invariant-graph}
We will now describe the GMT-GL graph construction for each model component, explaining which nodes are instantiated, which edges join them, and which parameter matrix, if any, sits on these edges. Throughout the construction, we use two recurring edge
connectivity patterns. The first captures any parameter matrix $\mathbf{M} \in \mathbb{R}^{m \times n}$ representing fully-connected layers of neurons $\mathcal{A} = \{a_i\}_{i=1}^m$ and $\mathcal{B} = \{b_i\}_{i=1}^n$. This is naturally instantiated as a \emph{full, bipartite} weighted edge set between the set of neuron nodes 
$\mathcal{A}$ and $\mathcal{B}$:
\begin{equation}
\label{eq:fc-edges}
\mathcal{E}_{\crossarrow}(\mathcal{A}, \mathcal{B}; \mathbf{M}) \coloneqq
\bigl\{ (a_i, b_j, \mathbf{M}_{i,j}) : 1 \le i \le m,\, 1 \le j \le n \bigr\},
\end{equation}
Under this setup, $\mathcal{A}, \mathcal{B}$ can be \emph{independently} permuted 
while maintaining the same graph structure, i.e. the permuted graph is isomorphic to the original.
Such permutations correspond to row and column permutations of the matrix $\mathbf{M}$,
which in turn provides the permutation invariance we seek.
The second pattern is a \emph{coupling} edge set between
two node-layers of equal size $m$, pairing same-index nodes without carrying any of the
transformer's trainable parameters:
\begin{equation}
\label{eq:coupling-edges}
\mathcal{E}_{\pararrow}(\mathcal{A}, \mathcal{B}) \;\coloneqq\; \bigl\{ (a_i, b_i) : 1 \le i \le m \bigr\}.
\end{equation}
This fixes the two layers to the same permutation. This is what lets us realize the couplings between permutations that
\Cref{tab:symmetries} requires,  $P_d$ across a block's sub-layers, and $P_V$ between the embedding and unembedding.
Moving forward, we now describe the graph construction for a single transformer block language model, which we also depict in~\Cref{fig:GL-symm-1-block}.

\textit{Embedding.} The graph starts with the node-layer
$\mathcal{V}_{V} = \{v_i\}_{i=1}^{|V|}$
(one node per vocabulary token), and the node-layer
$\mathcal{V}_{d} = \{x_i\}_{i=1}^{d}$ (one node per feature), joined by the fully-connected edge set
$\mathcal{E}_{\crossarrow}(\mathcal{V}_{V}, \mathcal{V}_{d}; \mathbf{E})$ carrying the embedding matrix.
Note that row and column permutation of $\mathbf{E}$ correspond to permuting $\mathcal{V}_{V}, \mathcal{V}_{d}$ which leaves the graph unchanged.

\textit{Multi-head attention.} first, to consider the pre-attention
normalization,
we introduce the node-layer $\mathcal{V}_{\gamma,\beta}^{a,\ell} =
\{g^{a,\ell}, s^{a,\ell}\}$, joined to $\mathcal{V}_d$ by the
fully-connected edge set $\mathcal{E}_{\crossarrow}(\mathcal{V}_{\gamma,\beta}^{a,\ell}, \mathcal{V}_d;
[\boldsymbol{\gamma}^{a,\ell}; \boldsymbol{\beta}^{a,\ell}])$.
Now, to respect the general-linear symmetry relating
$\mathbf{W}^{q,h}, \mathbf{W}^{k,h}$ and $\mathbf{W}^{v,h}, \mathbf{W}^{O}$, as anticipated, we work directly with the matrix products $\mathbf{W}^{q,h}(\mathbf{W}^{k,h})^\top$
and $\mathbf{W}^{v,h}\mathbf{W}^{O,h}$, where $\mathbf{W}^{O,h}$ is the $h$-th row-block of
$\mathbf{W}^{O}$.
Concretely, for every head $h = 1, \dots, H$ we introduce a node-layer $\mathcal{V}_{d}^{h} = \{x_i^h\}_{i=1}^d$ of $d$
nodes, joined to $\mathcal{V}_d$ by the edge set $\mathcal{E}_{\crossarrow}(\mathcal{V}_d,
\mathcal{V}_d^h; \mathbf{W}^{q,h}(\mathbf{W}^{k,h})^\top)$.
To additionally tie the column
permutation of the embedding to
the row permutation of $\mathbf{W}^{q,h}$, $\mathbf{W}^{k,h}$ and $\mathbf{W}^{v,h}$, we add, for
every head, the coupling edge set $\mathcal{E}_{\pararrow}(\mathcal{V}_d, \mathcal{V}_d^h)$.
However, these edges alone do not bind the $d$ nodes of a head \emph{together}: the nodes
$\{x_i^h\}_{h=1}^H$ of each index $i$ could be permuted across heads independently of the other
indices, which introduce larger graph symmetry than the one of the corresponding transformer.
To tie all nodes of
a head to the same head permutation, we introduce, similarly to \citet{knyazev2025accelerating}, for every head a single head node $u^h$, joined to
all nodes of that head by the parameter-free edge set $\{(u^h, x_i^h) : 1 \le i \le d\}$.
Next, each of these $H$ node-layers is joined to a
shared node-layer $\mathcal{V}_d' = \{x_i'\}_{i=1}^d$ of $d$ nodes by a fully-connected edge
set $\mathcal{E}_{\crossarrow}(\mathcal{V}_d^h, \mathcal{V}_d'; \mathbf{W}^{v,h}\mathbf{W}^{O,h})$, merging the
heads back into the residual stream.
Finally, to account for the residual connection,
we add the coupling edge set $\mathcal{E}_{\pararrow}(\mathcal{V}_d, \mathcal{V}_d')$
between the sub-layer's input and output layers.
As before, the pre-MLP normalization pair encoded with the node-layer $\mathcal{V}_{\gamma,\beta}^{b,\ell} = \{g^{b,\ell},
s^{b,\ell}\}$ joined to $\mathcal{V}_d'$ by the edge set
$\mathcal{E}_{\crossarrow}(\mathcal{V}_{\gamma,\beta}^{b,\ell}, \mathcal{V}_d'; [\boldsymbol{\gamma}^{b,\ell};
\boldsymbol{\beta}^{b,\ell}])$.

\textit{Feed-forward.} We continue with a
node-layer $\mathcal{V}_{ff} = \{z_i\}_{i=1}^{d_{\mathrm{ff}}}$ of $d_{\mathrm{ff}}$ nodes,
joined to $\mathcal{V}_d'$ by $\mathcal{E}_{\crossarrow}(\mathcal{V}_d',
\mathcal{V}_{ff}; \mathbf{W}^{1})$, and a further node-layer $\mathcal{V}_d'' = \{x_i''\}_{i=1}^d$
of $d$ nodes, joined to $\mathcal{V}_{ff}$ by $\mathcal{E}_{\crossarrow}(\mathcal{V}_{ff},
\mathcal{V}_d''; \mathbf{W}^{2})$; as above, coupling edge set $\mathcal{E}_{\pararrow}(\mathcal{V}_d',
\mathcal{V}_d'')$ accounts for residual connections. The node-layer
$\mathcal{V}_d''$ plays the role of $\mathcal{V}_d$ for the next block, and this pattern repeats for all blocks.

\textit{Unembedding.} After the last block, a final node-layer $\mathcal{V}_V' = \{v_i'\}_{i=1}^{|V|}$
is joined to the last $\mathcal{V}_d$ by the fully-connected edge set
$\mathcal{E}_{\crossarrow}(\mathcal{V}_d, \mathcal{V}_V'; \mathbf{F})$ carrying the unembedding matrix. To tie the
row permutation of the embedding matrix to the column permutation of the unembedding matrix -- both
equal to the shared vocabulary permutation $P_V$ -- we add the coupling edge set
$\mathcal{E}_{\pararrow}(\mathcal{V}_V, \mathcal{V}_V')$ between the network's very first and very last node
layers. The final normalization pair $\boldsymbol{\gamma}^{f}, \boldsymbol{\beta}^{f}$ is attached to
this last $\mathcal{V}_d$ in the same way, via the node-layer $\mathcal{V}_{\gamma,\beta}^{f} =
\{g^{f}, s^{f}\}$ joined to it by the fully-connected edge set
$\mathcal{E}_{\crossarrow}(\mathcal{V}_{\gamma,\beta}^{f}, \mathcal{V}_d; [\boldsymbol{\gamma}^{f};
\boldsymbol{\beta}^{f}])$.

\begin{figure}[H]
\centering
\colorlet{gmtRa}{teal!70}         
\colorlet{gmtU}{black!50}         
\colorlet{gmtQK}{red!90!black}    
\colorlet{gmtVO}{blue!50}         
\colorlet{gmtQ}{gray!60}          
\colorlet{gmtK}{orange!70}        
\colorlet{gmtV}{blue!50}          
\colorlet{gmtO}{violet!70}        
\resizebox{\linewidth}{!}{%
\begin{tikzpicture}[
    neuron/.style={circle, draw, minimum size=16pt, inner sep=0pt, fill=white, font=\tiny},
    layer/.style={neuron, fill=#1!14, draw=#1!60},
    qkv layer/.style={neuron, draw=black!45, top color=gmtQ!22, middle color=gmtK!22, bottom color=gmtV!22},
    panel/.style={draw=black!35, fill=black!3, rounded corners=6pt, line width=0.6pt},
    panel title/.style={font=\normalsize\bfseries, anchor=south east},
    >=Stealth,
    brace above/.style={decorate, decoration={brace, amplitude=6pt, mirror}},
]

\foreach \i in {1,...,4} {
    \node[layer=gmtRa] (A-R0-\i) at (0, -\i*0.9 - 2.3) {$r^{0}_{\i}$};
}
\foreach \i in {1,...,4} {
    \node[layer=gmtQK] (A-X1-\i) at (3.9, -\i*0.9 - 0.5) {$x^{1,1}_{\i}$};
    \node[layer=gmtQK] (A-X2-\i) at (3.9, -\i*0.9 - 4.1) {$x^{2,1}_{\i}$};
}
\foreach \i in {1,...,4} {
    \node[layer=gmtVO] (A-R1-\i) at (6.8, -\i*0.9 - 2.3) {$r^{1}_{\i}$};
}
\node[layer=gmtU, dashed] (A-U1) at (2.8, -0.45) {$u^{1,1}$};
\node[layer=gmtU, dashed] (A-U2) at (2.8, -8.55) {$u^{2,1}$};

\begin{pgfonlayer}{background}
\draw[panel] (-0.8, -9.15) rectangle (7.5, 0.85);
\foreach \i in {1,...,4} {
    \foreach \j in {1,...,4} {
        \draw[->, gmtQK] (A-R0-\i) -- (A-X1-\j);
        \draw[->, gmtQK] (A-R0-\i) -- (A-X2-\j);
        \draw[->, gmtVO] (A-X1-\i) -- (A-R1-\j);
        \draw[->, gmtVO] (A-X2-\i) -- (A-R1-\j);
    }
}
\foreach \i in {1,...,4} {
    \draw[->, thin, black] (A-R0-\i) to[bend left=30, looseness=1.3] (A-X1-\i);
    \draw[->, thin, black] (A-R0-\i) to[bend right=30, looseness=1.3] (A-X2-\i);
    \draw[->, thin, dashed, black] (A-U1) -- (A-X1-\i);
    \draw[->, thin, dashed, black] (A-U2) -- (A-X2-\i);
}
\foreach \i in {1,...,2} {
    \draw[->, thin, black] (A-R0-\i) .. controls (0.85, -0.4) and (5.1, 0.3) .. (A-R1-\i);
}
\foreach \i in {3,...,4} {
    \draw[->, thin, black] (A-R0-\i) .. controls (-0.25, -7.9) and (5.1, -9.2) .. (A-R1-\i);
}
\end{pgfonlayer}

\draw[brace above] (3.5, -1.4 + 0.15) -- (3.5, -4.1 - 0.15)
    node[midway, left=8pt, font=\footnotesize] {Head 1};
\draw[brace above] (3.5, -5.0 + 0.15) -- (3.5, -7.7 - 0.15)
    node[midway, left=8pt, font=\footnotesize] {Head 2};

\node[above, font=\small\bfseries, text=red!90!black] at (1.95, 0.0) {$W^{(Q,i)} {W^{(K,i)}}^\top$};
\node[above, font=\small\bfseries, text=blue!80!black] at (5.35, 0.0) {$W^{(V,i)} W^{(O,i)}$};
\node[panel title] at (7.35, -9.0) {GMT-GL (ours)};

\begin{scope}[xshift=8.7cm]
\foreach \i in {1,...,4} {
    \node[layer=gmtRa] (B-R0-\i) at (0, -\i*0.9 - 2.3) {$r^{0}_{\i}$};
}
\foreach \i in {1,...,2} {
    \node[layer=gmtQ] (B-Y1-\i) at (1.7, -\i*0.9 - 1.4) {$y^{1,1}_{\i}$};
    \node[layer=gmtQ] (B-Y2-\i) at (1.7, -\i*0.9 - 5.0) {$y^{2,1}_{\i}$};
}
\foreach \i in {1,...,4} {
    \node[layer=gmtK] (B-X1-\i) at (3.4, -\i*0.9 - 0.5) {$x^{1,1}_{\i}$};
    \node[layer=gmtK] (B-X2-\i) at (3.4, -\i*0.9 - 4.1) {$x^{2,1}_{\i}$};
}
\foreach \i in {1,...,2} {
    \node[layer=gmtV] (B-W1-\i) at (5.1, -\i*0.9 - 1.4) {$w^{1,1}_{\i}$};
    \node[layer=gmtV] (B-W2-\i) at (5.1, -\i*0.9 - 5.0) {$w^{2,1}_{\i}$};
}
\foreach \i in {1,...,4} {
    \node[layer=gmtO] (B-R1-\i) at (6.8, -\i*0.9 - 2.3) {$r^{1}_{\i}$};
}

\begin{pgfonlayer}{background}
\draw[panel] (-0.8, -9.15) rectangle (7.5, 0.85);
\foreach \i in {1,...,4} {
    \foreach \j in {1,...,2} {
        \draw[->, gmtQ] (B-R0-\i) -- (B-Y1-\j);
        \draw[->, gmtQ] (B-R0-\i) -- (B-Y2-\j);
        \draw[->, gmtK] (B-Y1-\j) -- (B-X1-\i);
        \draw[->, gmtK] (B-Y2-\j) -- (B-X2-\i);
        \draw[->, gmtV] (B-X1-\i) -- (B-W1-\j);
        \draw[->, gmtV] (B-X2-\i) -- (B-W2-\j);
        \draw[->, gmtO] (B-W1-\j) -- (B-R1-\i);
        \draw[->, gmtO] (B-W2-\j) -- (B-R1-\i);
    }
}
\foreach \i in {1,...,4} {
    \draw[->, thin, black!65] (B-R0-\i) .. controls +(1.1,0.9) and +(-1.1,0.3) .. (B-X1-\i);
    \draw[->, thin, black!65] (B-R0-\i) .. controls +(1.1,-0.9) and +(-1.1,-0.3) .. (B-X2-\i);
}
\foreach \i in {1,...,2} {
    \draw[->, thin, black!50] (B-R0-\i) .. controls (0.85, -0.4) and (5.1, 0.3) .. (B-R1-\i);
}
\foreach \i in {3,...,4} {
    \draw[->, thin, black!50] (B-R0-\i) .. controls (-0.25, -7.9) and (5.1, -9.2) .. (B-R1-\i);
}
\end{pgfonlayer}

\draw[brace above] (1.3, -2.3 + 0.15) -- (1.3, -3.2 - 0.15)
    node[midway, left=8pt, font=\footnotesize] {Head 1};
\draw[brace above] (1.3, -5.9 + 0.15) -- (1.3, -6.8 - 0.15)
    node[midway, left=8pt, font=\footnotesize] {Head 2};

\node[above, font=\small\bfseries, text=gray!90] at (0.85, 0.0) {$W^{(Q,i)}$};
\node[above, font=\small\bfseries, text=orange!80!black] at (2.55, 0.0) {${W^{(K,i)}}^\top$};
\node[above, font=\small\bfseries, text=blue!80!black] at (4.25, 0.0) {$W^{(V,i)}$};
\node[above, font=\small\bfseries, text=violet!80!black] at (5.95, 0.0) {$W^{(O,i)}$};
\node[panel title] at (7.35, -9.0) {GMT-P (ours)};
\end{scope}

\begin{scope}[xshift=17.4cm]
\foreach \i in {1,...,4} {
    \node[layer=gmtRa] (C-R0-\i) at (0, -\i*0.9 - 2.3) {};
}
\foreach \i in {1,...,2} {
    \node[qkv layer] (C-M1-\i) at (1.8, -\i*1.1 - 1.2) {};
    \node[qkv layer] (C-M2-\i) at (1.8, -\i*1.1 - 4.6) {};
}
\foreach \i in {1,...,4} {
    \node[layer=gmtO] (C-R1-\i) at (3.6, -\i*0.9 - 2.3) {};
}

\begin{pgfonlayer}{background}
\draw[panel] (-0.8, -9.15) rectangle (4.3, 0.85);
\foreach \i in {1,...,4} {
    \foreach \m in {M1,M2} {
        \foreach \j in {1,...,2} {
            \draw[->, gmtQ] (C-R0-\i) to[bend left=11] (C-\m-\j);
            \draw[->, gmtK] (C-R0-\i) -- (C-\m-\j);
            \draw[->, gmtV] (C-R0-\i) to[bend right=11] (C-\m-\j);
            \draw[->, gmtO] (C-\m-\j) -- (C-R1-\i);
        }
    }
}
\foreach \i in {1,...,2} {
    \draw[->, thin, black!50] (C-R0-\i) .. controls (0.45, -0.4) and (2.7, 0.3) .. (C-R1-\i);
}
\foreach \i in {3,...,4} {
    \draw[->, thin, black!50] (C-R0-\i) .. controls (-0.15, -7.9) and (2.7, -9.2) .. (C-R1-\i);
}
\end{pgfonlayer}

\draw[brace above] (1.4, -2.3 + 0.15) -- (1.4, -3.4 - 0.15)
    node[midway, left=8pt, font=\footnotesize] {Head 1};
\draw[brace above] (1.4, -5.7 + 0.15) -- (1.4, -6.8 - 0.15)
    node[midway, left=8pt, font=\footnotesize] {Head 2};

\node[above, font=\small\bfseries] at (0.85, 0.0)
    {\textcolor{gray!90}{$W^{(Q,i)}$}\,\textcolor{orange!80!black}{$W^{(K,i)}$}\,\textcolor{blue!80!black}{$W^{(V,i)}$}};
\node[above, font=\small\bfseries, text=violet!80!black] at (3.15, 0.0) {$W^{(O,i)}$};
\node[panel title, align=right] at (4.15, -9.0) {GMN};
\end{scope}

\end{tikzpicture}%
}
\caption{The multi-head attention subgraph under the three graph constructions, for an attention layer with
$d{=}4$ and $H{=}2$ heads ($d_k{=}d_v{=}2$); the rest of GMT-GL and GMT-P is identical.
\textbf{Left:} GMT-GL places the products $W^{(Q,i)}{W^{(K,i)}}^\top$ and $W^{(V,i)}W^{(O,i)}$ on the edges,
and the head nodes $u^{h,1}$ bind the nodes of each head together.
\textbf{Middle:} GMT-P gives each projection matrix its own edge set.
\textbf{Right:} \cite{lim2024graph} and \cite{kofinas2024graph} which share the same construction for MHA, noted as GMN.}
\label{fig:mha-graph-comparison}
\end{figure}
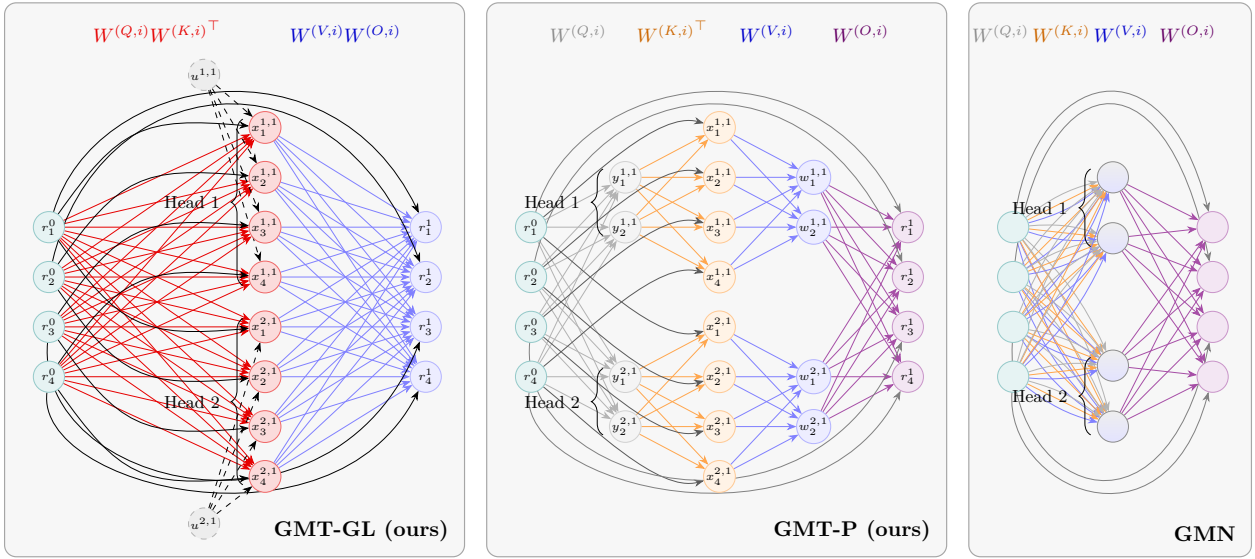

\textbf{Limitations of GMT-GL.} 
Our construction \emph{GMT-GL}, is equivariant to the whole group $\mathcal{G}$,
particularly including the general-linear symmetry of the attention mechanism.
This exact equivariance is of course desirable, but it
comes with two disadvantages. First, to respect the general-linear symmetry, $\mathbf{W}^{q,h}$ and
$\mathbf{W}^{k,h}$ --- each of shape $d \times d_{k}$ --- are multiplied together to form a single
matrix of shape $d \times d$ placed on the graph's edges. The size of this product is $d^{2}$, and in the typical case where $d_{k} \ll d$, this is much larger than the combined size of the two matrices it replaces, that is, $2 d d_{k}$. As a result, the graph has far more edges than the layer has parameters. The same holds for the general-linear symmetry relating $\mathbf{W}^{v,h}$ and $\mathbf{W}^{O,h}$, with product
$\mathbf{W}^{v,h}\mathbf{W}^{O,h}$.Second, since GMT-GL edges carry entries of matrix products rather than individual weights,
model weights no longer correspond to graph edges, which make weight-level tasks, where
a property of each weight must be predicted, more challenging.

\textbf{Permutation-Invariant Graph Construction (GMT-P).}
\label{sec:perm-invariant-graph}
To mitigate the aforementioned issues with GMT-GL, we propose a more efficient alternative, \emph{GMT-P}, which
restricts the symmetry group to permutation matrices alone, $\mathcal{P}_{d_k} \leq \mathrm{GL}_{d_{k}}, \mathcal{P}_{d_v} \leq \mathrm{GL}_{d_{v}}$, and works directly with the weight matrices
$\mathbf{W}^{q,h}$,
$\mathbf{W}^{k,h}$, $\mathbf{W}^{v,h}, \mathbf{W}^{O,h}$ themselves, rather than the two products.
Concretely, the GMT-P construction considers the subgroup $\tilde{\mathcal{G}} \leq \mathcal{G}$
defined by
\begin{equation}
  \tilde{\mathcal{G}} \;=\; \underbrace{S_{V} \times S_{d}}_{\text{Embedding}}
   \bigtimes_{\ell=1}^{L}
  \Bigl[\, \underbrace{\bigl(\mathcal{P}_{d_{k}} \times \mathcal{P}_{d_{v}}\bigr) \wr S_{H}}_{\text{attention}}
  \;\times\; \underbrace{S_{d_{\mathrm{ff}}}}_{\text{FF}} \Bigr],
  \label{eq:full-group-permutation}
\end{equation}

The full construction of a single-block transformer language
model under this second method is shown in \Cref{fig:P-symm-1-block}  (\Cref{app:graph-construction-figures}), and we give its formal definition in \Cref{app:graph-construction-gmt-p}. Each parameter now
corresponds directly to a single edge: GMT-P thus avoids both disadvantages of GMT-GL described
above: it is more compact, and a weight-level task on the model is equivalent to an edge-level task on the resulting graph.
On the transformer blocks, GMT-P encodes the same permutation symmetries as the construction of \citet{knyazev2025accelerating}, though through a different graph;
The two also differ in the embedding and unembedding, whose vocabulary permutation symmetry is not encoded by the latter.

\textbf{Node and edge features.}
For both constructions, each node is attributed with features encoding the index of the node-layer\footnote{Not to be confused with the transformer layer index $\ell$.} it belongs to
(for instance, the graph of \Cref{fig:GL-symm-1-block} has 7 node-layers)
together with its type, i.e.\ the network component it originates from: MHA head, MLP hidden unit etc.
Each edge has features encoding its type, i.e., weight edge or a coupling edge,
and its weight (if any). These encodings are necessary to  break unwanted symmetries of the graph
like permutation of nodes across different node-layers or
types, which do not correspond to the symmetry of the original transformer.

\begin{theorem}[Symmetries as graph isomorphisms]
\label{thm:graph-isomorphism}
Let $\Gamma_{\mathrm{GL}}(w)$ and $\Gamma_{\mathrm{P}}(w)$ denote the graphs
 that GMT-GL and GMT-P respectively produce from weights $w \in
\mathcal{W}$. Then for any $w, w' \in \mathcal{W}$,
\begin{align*}
\Gamma_{\mathrm{GL}}(w) \cong \Gamma_{\mathrm{GL}}(w') &\iff w' \in \mathcal{G} \cdot w, \\
\Gamma_{\mathrm{P}}(w) \cong \Gamma_{\mathrm{P}}(w') &\iff w' \in \tilde{\mathcal{G}} \cdot w,
\end{align*}
where $\mathcal{G} \cdot w$ and $\tilde{\mathcal{G}} \cdot w$ denote the orbits of $w$ under
$\mathcal{G}$ (\Cref{eq:full-group}) and $\tilde{\mathcal{G}}$
(\Cref{eq:full-group-permutation}) respectively, and $\cong$ denotes isomorphism of attributed
graphs (i.e.\ a bijection of nodes preserving node features, edges, and edge features).
For $\Gamma_{\mathrm{GL}}$, the implication ($\Rightarrow$) assumes that the per-head matrices
of $w$ have full column rank.
\end{theorem}
\Cref{thm:graph-isomorphism}, which we prove in \Cref{app:graph-construction-proof},
states that our graphs represent the weights faithfully up to
symmetry. The ($\Leftarrow$) direction establishes \emph{invariance}: weights related by a
symmetry yield isomorphic graphs, so any message-passing GNN is automatically invariant to
$\mathcal{G}$ (respectively $\tilde{\mathcal{G}}$). 
The ($\Rightarrow$) direction establishes \emph{separation}: weights not
related by a symmetry yield non-isomorphic graphs, so the construction discards no information
beyond the symmetry itself. 
Note that the theorem concerns the graph representation, not the GNN:
whether a particular GNN separates two non-isomorphic graphs depends on its expressive power
\citep{xu2019powerful}.

\paragraph{Metanetwork architecture}
\label{sec:method-architecture}
In practice, We take our graphs and feed them to a GNN.
As the backbone, we employ GINE layers \citep{hu2020strategies}, extension of GIN \citep{xu2019powerful} that incorporates edge features into the
message-passing procedure along with updates to the edge features.
Details on the architecture are in \Cref{app:experimental-details-architectures}.

\section{Experiments}
\label{sec:experiments}

We evaluate our approach on three tasks, instantiated over two new datasets of transformer language models. These gather populations differing in architecture,
training mixture, and fine-tuned components (see \Cref{app:experimental-details-datasets}). We will release our dataset to support future research.

\textbf{Baselines.} We compare GMT-GL and GMT-P against three families of baselines:
(i) the graph construction of \citet{lim2024graph} and \citet{kofinas2024graph},
which treat multi-head attention identically and are thus reported as a single baseline,
`GMN' (we do not separately evaluate the construction of \citet{knyazev2025accelerating}, as on the transformer blocks it encodes the same symmetries as GMT-P, see \Cref{sec:perm-invariant-graph}); (ii) the equivariant metanetwork of \citet{tran2025transformer}, denoted `NFN-T', which captures the general-linear symmetries but is tied to a single, fixed transformer architecture, so we compare against it only on a sub-population of models sharing one architecture; and (iii) two baselines that discard the transformer structure and operate on the flattened parameter vector: an MLP, also trained with data augmentations
(MLP+Aug), and DeepSets
\citep{zaheer2017deep}, which treats the parameters as a permutation-invariant set. See \Cref{app:experimental-details-architectures} for more details on architectures and baselines.

\subsection{Corpus-Mixture Prediction}\label{sec:exp-distribution}

\textbf{Setup.} Our first task is to predict, from a model's weights, the proportions of
the corpora it was trained on. We use a population of small language models
with $90$K--$150$K parameters, each trained on a mixture of three corpora, Shakespeare
\citep{karpathy2015charrnn}, TinyStories \citep{eldan2023tinystories}, and Wikipedia
\citep{merity2017pointer}, drawn at model-specific proportions, e.g.,
$(80\%,15\%,5\%)$. This constitutes the prediction target, and is a property of the whole model, \emph{invariant} to the weight-space symmetry group
$\mathcal{G}$ (\Cref{eq:full-group}). We pose the task as a regression onto the
probability simplex and report the KL divergence and MAE between the predicted and true
mixtures.

\begin{table}[t]
\caption{Performance of all methods on the corpus-mixture prediction task
(\Cref{sec:exp-distribution}).}
\label{tab:exp-distribution}
\begin{center}
\footnotesize
\setlength{\tabcolsep}{4pt}
\begin{tabular}{@{}lcc@{}}
\toprule
\multicolumn{1}{c}{\bf Method} & \multicolumn{1}{c}{\bf KL} &
\multicolumn{1}{c}{\bf MAE} \\
\midrule
\multicolumn{3}{l}{\emph{Flat weight vector baselines}} \\
DeepSets               & $0.2681 \pm 0.0002$ & $0.1974 \pm 0.0002$ \\
MLP                    & $0.2637 \pm 0.0067$ & $0.1951 \pm 0.0031$ \\
MLP+Aug                & $0.2715 \pm 0.0009$ & $0.1988 \pm 0.0003$ \\
\midrule
\multicolumn{3}{l}{\emph{Graph constructions}} \\
GMN                    & $0.1695 \pm 0.0123$ & $0.1439 \pm 0.0064$ \\
GMT-P (ours)                 & $0.1238 \pm 0.0058$ & $0.1210 \pm 0.0044$ \\
GMT-GL  (ours)               & $\mathbf{0.1236} \pm 0.0136$ & $\mathbf{0.1194} \pm 0.0082$ \\
\bottomrule
\end{tabular}
\end{center}
\end{table}

\begin{table}[t]
\caption{Performance comparison against NFN-T~\cite{tran2025transformer} over a subset of $500$ homogeneous architectures from the corpus-mixture dataset (\Cref{sec:exp-distribution}).}
\label{tab:exp-distribution-tran}
\begin{center}
\footnotesize
\setlength{\tabcolsep}{4pt}
\begin{tabular}{@{}lcc@{}}
\toprule
\multicolumn{1}{c}{\bf Method} & \multicolumn{1}{c}{\bf KL} &
\multicolumn{1}{c}{\bf MAE} \\
\midrule
NFN-T                    & $0.2942 \pm 0.0060$ & $0.2101 \pm 0.0008$ \\
GMT-P   (ours)         & $0.2040 \pm 0.0121$ & $\mathbf{0.1644} \pm 0.0080$ \\
GMT-GL  (ours)         & $\mathbf{0.2038} \pm 0.0253$ & $0.1652 \pm 0.0126$ \\
\bottomrule
\end{tabular}
\end{center}
\end{table}

\begin{figure}[H]
    \centering
    \includegraphics[width=0.6\linewidth]{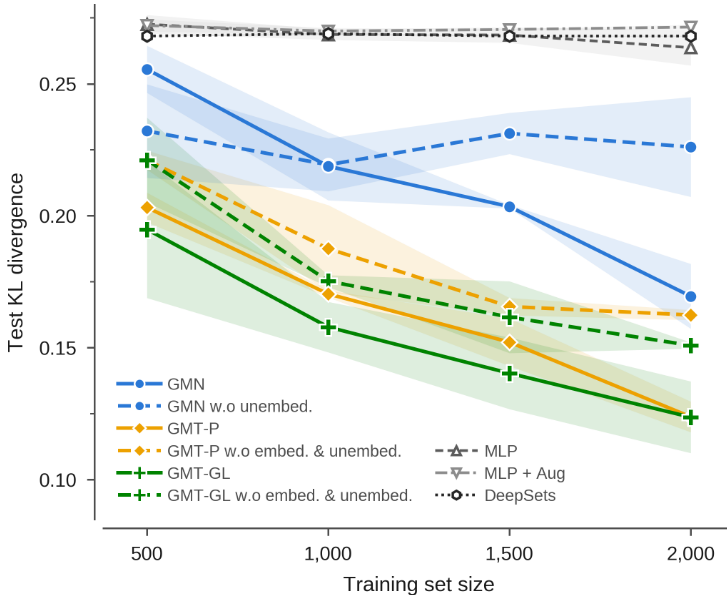}
    \caption{Test KL divergence on corpus-mixture prediction, evaluated as a
    function of training set size and across ablations of the (un)embedding components.}
    \label{fig:corpus-mixture-kl-compact}
\end{figure}

\textbf{Results} in \Cref{tab:exp-distribution} show a wide gap between the flat-weight
baselines and the graph-based models: DeepSets and MLP are largely outperformed, even
when augmentations are used to compensate for the MLP's lack of invariance (see~\Cref{app:experimental-details-architectures}). Among
graph-based models, the two GMT variants perform similarly and clearly outperform GMN;
this gap holds across metrics and quantity of available training data (see~\Cref{fig:corpus-mixture-kl-compact} and paragraph below). On the fixed-architecture sub-population
(\Cref{tab:exp-distribution-tran}), both variants also substantially outperform NFN-T.

\textbf{Ablating embedding and unembedding.} GMN encodes the final transformer classification layer but, unlike GMT, not the token embedding. To test whether this explains the performance gap, and to what extent the embedding and unembedding modules are predictive for the pretraining mix, here, we ablate both this two components from all inputs. We additionally vary the training set size to assess sample efficiency and report resulst in~\Cref{fig:corpus-mixture-kl-compact}. The ablation degrades all models, confirming that (un)embedding components carry a non-negligible part of the predictive signal. GMT still clearly outperforms GMN in both settings and across data regimes; in fact, without the unembedding, GMN no longer benefits from more training data (dashed blue line). This confirms the importance of the additional inductive biases captured by our
constructions.

\subsection{Fine-tuning Detection}
\label{sec:exp-finetune-classification}

\textbf{Setup.} Our task deals with detecting whether a model has been finetuned. We construct a second dataset consisting of $4{,}000$ transformer language models with 3M--5M parameters, all trained on TinyShakespeare \citep{karpathy2015charrnn}; half are subsequently LoRA fine-tuned on Wikipedia~\citep{merity2017pointer}. To make the problem harder and the population more diverse, fine-tuning adapts only a random subset of each model's weight matrices. Here the label is attached to the whole model, so this is again an invariant task. As customary, the embedding and unembedding modules are not fine-tuned, so we omit their parameters and corresponding graph components in these experiments.

\textbf{Results} are reported in~\Cref{tab:exp-finetune-classification} in terms of accuracy, macro-F1 and ROC-AUC. The MLP and DeepSets baselines still largely underperform. On the contrary, the graph-aware models generally obtain strong performance, with our GMT models outperforming GMN. Best performance is obtained, in particular, by the GMT-GL variant.

\begin{table}[t]
\caption{Results on fine-tuning detection (\Cref{sec:exp-finetune-classification}).}
\label{tab:exp-finetune-classification}
\begin{center}
\small
\begin{tabular}{@{}lccc@{}}
\toprule
\multicolumn{1}{c}{\bf Method} & \multicolumn{1}{c}{\bf ROC-AUC} &
\multicolumn{1}{c}{\bf Macro-F1} & \multicolumn{1}{c}{\bf Accuracy} \\
\midrule
MLP                                     & $0.6237$ & $0.3333$ & $0.5$ \\
MLP+Aug                                 & $0.5891$ & $0.3333$ & $0.5$ \\
DeepSets                                & $0.5777$ & $0.3333$ & $0.5$ \\
\midrule
GMN                                     & $0.9359$ & $0.8370$ & $0.8374$ \\
GMT-P (ours)                            & $0.9495$ & $0.8473$ & $0.8475$ \\
GMT-GL (ours)                           & $\mathbf{0.9557}$ & $\mathbf{0.8611}$ & $\mathbf{0.8613}$ \\
\bottomrule
\end{tabular}
\end{center}
\end{table}
 
\subsection{Weight-level Fine-tuning Detection}
\label{sec:exp-weight-classification}

\textbf{Setup.} Our third task asks not only whether a model was fine-tuned, but which of its weights were updated. We reuse the fine-tuned models of \Cref{sec:exp-finetune-classification}: since only a random subset of their weight matrices was adapted, each contains both updated and untouched weights. The target is now attached to each parameter rather than to the whole model, making this an \emph{equivariant} task. In GMT-P every weight is a single edge, so the task becomes edge-level binary classification, and GMT-GL is excluded, as its edges do not always correspond to individual weights. Flat-weight baselines are adapted to predict one label per parameter. As most weights are untouched, we report macro-F1 as primary metric, alongside accuracy and ROC-AUC.

\textbf{Results} are reported in~\Cref{tab:exp-weight-classification}, and show that our GMT-P model outperforms all other baselines, marking an absolute gap of 5\% in terms of Macro-F1 w.r.t. the GMN runnerup. The flat-weight baselines again trail the graph-based models by a wide margin.

\begin{table}[t]
\caption{Weight-lv fine-tuning detection (\Cref{sec:exp-weight-classification}).}
\label{tab:exp-weight-classification}
\begin{center}
\small
\begin{tabular}{@{}lccc@{}}
\toprule
\multicolumn{1}{c}{\bf Method} &
\multicolumn{1}{c}{\bf Macro-F1} & \multicolumn{1}{c}{\bf ROC-AUC} & \multicolumn{1}{c}{\bf Accuracy} \\
\midrule
MLP                                     & 0.4314 & 0.5606 & 0.7587 \\
MLP+Aug                                 & 0.5260 & 0.7797 & 0.7593 \\
DeepSets                                & 0.4313 & 0.5502 & 0.7586 \\
\midrule
GMN                                     & $0.9043$ & $0.9743$ & $0.9043$ \\
GMT-P (ours)                            & $\mathbf{0.9520}$ & $\mathbf{0.9926}$ & $\mathbf{0.9520}$ \\
\bottomrule
\end{tabular}
\end{center}
\end{table}

\section{Conclusions}

We proposed GMT, a graph metanetwork for transformer language models. We formalized a symmetry group of a \emph{complete} decoder-only model, combining residual-stream, feed-forward, head, vocabulary permutations and general-linear symmetries of attention. We proposed a graph construction which uses parameter-free coupling edges and head nodes so that, unlike earlier approaches, its isomorphisms match the aforementioned group: we showed that, under mild rank conditions, two models yield isomorphic graphs iff they are related by one such group element. A standard GNN is thus invariant  to it by construction, for any depth, width, and number of heads. We proposed two variants: GMT-GL captures the full group, while GMT-P restricts the general-linear symmetries to permutations in exchange for a more compact graph with one edge per parameter, also enabling weight-level tasks. On two new populations of language models, GMT outperforms existing baselines on corpus-mixture prediction and on model- and weight-level fine-tuning detection, and its advantage over prior graph metanetworks persists across training set sizes.

\textbf{Limitations and future work.} $\mathcal{G}$ is not maximal: e.g.\ scaling and sign symmetries of some activations are left untreated and could be handled. Our experiments are limited to models of a few million parameters; scaling to much larger models will likely require tailored graph learning approaches. Finally, our tasks are small-scale proxies for auditing questions; we hope our datasets encourage progress towards provenance and modification analysis of real-world open-weight models, as well as other equivariant applications such as weight editing.

\subsection*{Acknowledgements}
Guy Bar-Shalom is supported by the Jacobs Qualcomm PhD Fellowship. Fabrizio Frasca conducted this work supported by an Aly Kaufman Post-Doctoral Fellowship. Haggai Maron is supported by the Israel Science Foundation through a personal grant (ISF 264/23)
and an equipment grant (ISF 532/23), and by the Career Advancement Chairs in Artificial
Intelligence - Schmidt Futures.

\bibliographystyle{plainnat}
\bibliography{references}

\clearpage
\appendix
\section{Extended related work} \label{app:ext_prev}

\textbf{Equivariant neural networks.}
Designing neural architectures that are invariant or equivariant to input symmetries is a central theme
of geometric deep learning \citep{bronstein2021geometric}, with examples ranging from
group-equivariant and steerable CNNs
\citep{cohen2016group,weiler20183d}
and rotation-equivariant models
\citep{cohen2018spherical,esteves2018learning,thomas2018tensor}
to general constructions
\citep{ravanbakhsh2017equivariance,kondor2018generalization,finzi2021practical}.
Most relevant to weight spaces are permutation-equivariant models for sets
\citep{zaheer2017deep,qi2017pointnet,hartford2018deep,maron2020learning} and graphs
\citep{maron2019invariant,maron2019provably}.

\textbf{Weight-space learning.}
Neural network weights have long been known to admit symmetries
\citep{hecht1990algebraic,godfrey2022symmetries}, which have been exploited for model
alignment and merging \citep{entezari2021role, ainsworth2022git}. Following early
symmetry-agnostic methods
\citep{unterthiner2020predicting,eilertsen2020classifying,schurholt2021selfsupervised,schurholt2024saneicml},
equivariant metanetworks were designed for the neuron permutations of MLPs and CNNs
\citep{navon2023equivariant,zhou2023permutation,zhou2023nft}, and extended to arbitrary
permutation-symmetric weight spaces \citep{zhou2024universal} and to scaling and sign symmetries
\citep{tran2024monomial,vo2025polynomial,kalogeropoulos2024scalegmn}; see
\citet{han2026weightspace} for a survey. Graph metanetworks \citep{lim2024graph,kofinas2024graph},
which we build on, encode networks as graphs processed by GNNs.

\textbf{Transformer weight spaces.}
The general-linear symmetries of attention have been studied for linear mode connectivity
\citep{theus2025generalized}, and analogous symmetries for processing low-rank adapters
\citep{putterman2024learning}. \citet{tran2025transformer} design layers equivariant to the
attention symmetries, but for a single, fixed architecture. Previously proposed graph metanetworks encode attention
layers without the general-linear symmetries.
\citet{lim2024graph} suggest, but do not use, head-index features, which would, either way, violate the
head-permutation symmetry; \citet{kofinas2024graph} share nodes across the query, key, and
value projections of all heads, tying the query--key and value--output
permutations and admitting spurious exchanges of dimensions across heads.
\citet{knyazev2025accelerating} address the latter issues in the context of accelerating training
by predicting future weights: they add a node per head, connected to its query--key and value
dimensions, and keep separate nodes for the query--key and value--output dimensions, thereby
capturing, on the transformer blocks, the same permutation symmetries as our GMT-P.
They include the token embedding in the graph, but do not encode its vocabulary permutation
symmetry: when models share a tokenizer, vocabulary nodes are given token-index features,
breaking it, and the unembedding is never coupled to the embedding.
None of these works addresses the symmetries of a complete language model, including the
vocabulary permutation coupling its embedding and unembedding.

\section{Theorems proofs}
\label{app:proofs}

\subsection{Proof of \texorpdfstring{\Cref{thm:symmetries}}{Theorem 1}}
\label{app:symmetries-proof}

Throughout this section we fix the architecture of \Cref{sec:method-symmetries}:
decoder-only transformer with $L$ blocks, $H$ heads per block, embedding dimension $d$, head
dimensions $d_{k}, d_{v}$, feed-forward hidden width $d_{\mathrm{ff}}$ and vocabulary size $V$.

\textbf{Notation and conventions.}
For $n \in \mathbb{N}$ we write $[n] = \{1, \dots, n\}$, we
identify a permutation $\pi \in S_{n}$ with the permutation matrix $P_{\pi} \in \{0,1\}^{n \times
n}$ and we write $\Pi_{n} \subset
\mathbb{R}^{n \times n}$ for the set of such matrices.
We denote $\mathbf{1} \in \mathbb{R}^{n}$ as the all-ones vector.
We write $\odot$ for the entrywise product,
$\phi: \mathbb{R} \to \mathbb{R}$ for the pointwise nonlinearity of the feed-forward sub-layer.

\textbf{The forward pass.}
Let $\mathbf{t} = (t_{1}, \dots, t_{N}) \in [V]^{N}$ be a sequence of $N$ tokens and let
$\mathbf{T} = \mathbf{T}(\mathbf{t}) \in \{0,1\}^{N \times V}$ be its one-hot matrix, where
$\mathbf{T}_{n v} = \delta_{v, t_{n}}$. Given weights $w \in \mathcal{W}$,
the network computes the following sequence of activations. Defining the norm
applied to each row $\mathbf{x} \in \mathbb{R}^{1 \times d}$ as
$\mathrm{Norm}(\mathbf{x}; \boldsymbol{\gamma}, \boldsymbol{\beta}) = \frac{\mathbf{x} -
\mu(\mathbf{x})\mathbf{1}^{\top}}{\sigma(\mathbf{x})} \odot \boldsymbol{\gamma} +
\boldsymbol{\beta}$, we have
\begin{align}
\label{eq:app-fwd-emb}
\text{(1) embedding:} &&
\mathbf{X}^{0} &= \mathbf{T}\,\mathbf{E} \;\in\; \mathbb{R}^{N \times d},
\\[2pt]
\label{eq:app-fwd-norma}
\text{(2) pre-attn.\ norm:} &&
\tilde{\mathbf{X}}^{\ell-1} &= \mathrm{Norm}\bigl(\mathbf{X}^{\ell-1};
\boldsymbol{\gamma}^{a,\ell}, \boldsymbol{\beta}^{a,\ell}\bigr) \;\in\; \mathbb{R}^{N \times d},
\\[2pt]
\label{eq:app-fwd-logits}
\text{(3a) attention scores:} &&
\mathbf{S}^{h,\ell} &= \frac{\tilde{\mathbf{X}}^{\ell-1} \mathbf{W}^{q,h,\ell} (\mathbf{W}^{k,h,\ell})^{\top}
(\tilde{\mathbf{X}}^{\ell-1})^{\top}}{\sqrt{d_{k}}} \;\in\; \mathbb{R}^{N \times N},
\\[2pt]
\label{eq:app-fwd-attn}
\text{(3b) attention:} &&
\mathbf{Y}^{\ell} &= \mathbf{X}^{\ell-1} + \sum_{h=1}^{H}
\softmax\bigl(\mathbf{S}^{h,\ell}\bigr)\, \tilde{\mathbf{X}}^{\ell-1}
\mathbf{W}^{v,h,\ell} \mathbf{W}^{O,\ell}_{h},
\\[2pt]
\label{eq:app-fwd-normb}
\text{(4) pre-FF norm:} &&
\tilde{\mathbf{Y}}^{\ell} &= \mathrm{Norm}\bigl(\mathbf{Y}^{\ell};
\boldsymbol{\gamma}^{b,\ell}, \boldsymbol{\beta}^{b,\ell}\bigr) \;\in\; \mathbb{R}^{N \times d},
\\[2pt]
\label{eq:app-fwd-ff}
\text{(5) feed-forward:} &&
\mathbf{X}^{\ell} &= \mathbf{Y}^{\ell} +
\phi\bigl(\tilde{\mathbf{Y}}^{\ell} \mathbf{W}^{1,\ell}\bigr) \mathbf{W}^{2,\ell},
\\[2pt]
\label{eq:app-fwd-normf}
\text{(6) final norm:} &&
\tilde{\mathbf{X}}^{L} &= \mathrm{Norm}\bigl(\mathbf{X}^{L};
\boldsymbol{\gamma}^{f}, \boldsymbol{\beta}^{f}\bigr) \;\in\; \mathbb{R}^{N \times d},
\\[2pt]
\label{eq:app-fwd-unemb}
\text{(7) unembedding:} &&
\mathbf{Z} &= \tilde{\mathbf{X}}^{L}\, \mathbf{F} \;\in\; \mathbb{R}^{N \times V},
\qquad
f_{w}(\mathbf{t}) = \softmax(\mathbf{Z}),
\end{align}
for $\ell = 1, \dots, L$, where \Cref{eq:app-fwd-norma,eq:app-fwd-logits,eq:app-fwd-attn,eq:app-fwd-normb,eq:app-fwd-ff}
are iterated in this order, $\softmax$ is applied to each row separately.
Thus $f_{w} : [V]^{N} \to \mathbb{R}^{N \times V}$
maps a token sequence to the $N$ next-token distributions it induces,
one per position; the usual next-token distribution is its last row.

\textbf{Proof of the theorem.}

Let $w \in \mathcal{W}, s \in \mathcal{G}$, and let $w' = s \cdot w$ with the group
action given by \Cref{tab:symmetries}.
Fix a token sequence $\mathbf{t} \in [V]^{N}$ of arbitrary length $N$ and let $\mathbf{t}' =
\pi_{V}(\mathbf{t}) \in [V]^{N}$, be the same sequence read in the
relabeled dictionary according to the embedding row permutation of the embedding matrix $\mathbf{E}$.
We denote unprimed symbols as the activations
\Cref{eq:app-fwd-emb,eq:app-fwd-norma,eq:app-fwd-logits,eq:app-fwd-attn,eq:app-fwd-normb,eq:app-fwd-ff,eq:app-fwd-normf,eq:app-fwd-unemb}
produced by $w$ on $\mathbf{t}$, and primed symbols those produced by $w'$ on $\mathbf{t}'$.
We walk through the forward pass produced by $w'$ and show it outputs the same next-token distributions as
$w$, up to the permutation of the vocabulary dictionary.

\emph{(1) Embedding.} Notice the embedded vectors produced by $w'$ equals the embedded vectors
produced by $w$ after applying the column permutation $P_{d}$,
which will be used as input to the next sub-layer: 
\begin{equation}
\label{eq:app-step-emb}
{\mathbf{X}^{0}}' = \mathbf{T}(\mathbf{t}')\, \mathbf{E}'
= \mathbf{T}(\mathbf{t}) P_{V}P_{V}^{\top} \mathbf{E} P_{d}
= \mathbf{X}^{0} P_{d}.
\end{equation}

\emph{(2) Pre-attention normalization.} Since $\mathrm{Norm}$ acts on each row separately, we get:
\begin{equation}
\label{eq:app-step-norma}
{\tilde{\mathbf{X}}^{\ell-1}{}}'
= \mathrm{Norm}\bigl(\mathbf{X}^{\ell-1}P_{d};\, \boldsymbol{\gamma}^{a,\ell}P_{d},\,
  \boldsymbol{\beta}^{a,\ell}P_{d}\bigr)
= \frac{\mathbf{X}^{\ell-1}P_{d} - \mu\mathbf{1}^{\top}P_{d}}{\sigma}
  \odot \bigl(\boldsymbol{\gamma}^{a,\ell}P_{d}\bigr) + \boldsymbol{\beta}^{a,\ell}P_{d}
= \tilde{\mathbf{X}}^{\ell-1} P_{d},
\end{equation}
Which is true since $\mu, \sigma$ are permutation invariant,
therefore unchanged when computed on $\mathbf{X}^{\ell-1}P_{d}$ rather than $\mathbf{X}^{\ell-1}$.

\emph{(3a) Attention scores.} We write $\bar{h} = \sigma_{\ell}(h)$ (the head permutation of $s$)
and abbreviate $G_{qk} = G_{qk}^{\bar{h},\ell}$, $G_{vo} = G_{vo}^{\bar{h},\ell}$.
We get for each head that the attention logits are:
\begin{equation}
\label{eq:app-step-scores}
\begin{aligned}
{\mathbf{S}^{h,\ell}}'
&= \frac{{\tilde{\mathbf{X}}^{\ell-1}{}}'\, {\mathbf{W}^{q,h,\ell}}'
  ({\mathbf{W}^{k,h,\ell}}')^{\top}\, ({\tilde{\mathbf{X}}^{\ell-1}{}}')^{\top}}{\sqrt{d_{k}}}
\\[2pt]
&= \frac{(\tilde{\mathbf{X}}^{\ell-1}
  P_{d})(P_{d}^{\top}
  \mathbf{W}^{q,\bar{h},\ell}
  G_{qk})(G_{qk}^{-1}
  (\mathbf{W}^{k,\bar{h},\ell})^{\top}
  P_{d})(P_{d}^{\top}
  (\tilde{\mathbf{X}}^{\ell-1})^{\top})}{\sqrt{d_{k}}}
\;=\; \mathbf{S}^{\bar{h},\ell}.
\end{aligned}
\end{equation}
Meaning the attention pattern of slot $h$ in the transformed network is thus, entry for entry, that of head
$\bar{h}$ in the original one.

\emph{(3b) Attention.} Here we have the exact same cancelation between the $G_{vo}$ and $G_{qk}$ matrices,
and between the $P_{d}$ and $P_{d}^{\top}$ matrices, which gives
${\tilde{\mathbf{X}}^{\ell-1}{}}'{\mathbf{W}^{v,h,\ell}}'{\mathbf{W}^{O,\ell}_{h}}' =
\tilde{\mathbf{X}}^{\ell-1}\mathbf{W}^{v,\bar{h},\ell}\mathbf{W}^{O,\ell}_{\bar{h}}P_{d}$,
therefore we get that the output of the attention sub-layer is exactly:
\begin{equation}
\label{eq:app-step-attn}
{\mathbf{Y}^{\ell}}'
= \mathbf{X}^{\ell-1}P_{d} + \sum_{h=1}^{H} \softmax\bigl(\mathbf{S}^{\bar{h},\ell}\bigr)\,
  \tilde{\mathbf{X}}^{\ell-1} \mathbf{W}^{v,\bar{h},\ell} \mathbf{W}^{O,\ell}_{\bar{h}} P_{d}
\;=\; \mathbf{Y}^{\ell} P_{d}.
\end{equation}

\emph{(4) Pre-feed-forward normalization.} Identical to step (2), applied now to the output of
\Cref{eq:app-step-attn}:
\begin{equation}
\label{eq:app-step-norm}
{\tilde{\mathbf{Y}}^{\ell}{}}' = \mathrm{Norm}\bigl(\mathbf{Y}^{\ell}P_{d};
\boldsymbol{\gamma}^{b,\ell}P_{d}, \boldsymbol{\beta}^{b,\ell}P_{d}\bigr) =
\tilde{\mathbf{Y}}^{\ell}P_{d}
\end{equation}

\emph{(5) Feed-forward.} since $\phi$ (the pointwise non-linearity) commutes with
$\hat{P}^{\ell}$ we can substitute the attention output into \Cref{eq:app-fwd-ff} to get:
\begin{equation}
\label{eq:app-step-ff}
\begin{aligned}
{\mathbf{X}^{\ell}}'
&= {\mathbf{Y}^{\ell}}' + \phi\bigl({\tilde{\mathbf{Y}}^{\ell}{}}'\, {\mathbf{W}^{1,\ell}}'\bigr)
  {\mathbf{W}^{2,\ell}}'
\\[2pt]
&= \mathbf{Y}^{\ell}P_{d}
  + \phi\bigl(\tilde{\mathbf{Y}}^{\ell} P_{d} P_{d}^{\top} \mathbf{W}^{1,\ell} \hat{P}^{\ell}\bigr)
  (\hat{P}^{\ell})^{\top} \mathbf{W}^{2,\ell} P_{d}
\\[2pt]
&= \mathbf{Y}^{\ell}P_{d} + \phi\bigl(\tilde{\mathbf{Y}}^{\ell}\mathbf{W}^{1,\ell}\bigr)
  \hat{P}^{\ell}(\hat{P}^{\ell})^{\top}
  \mathbf{W}^{2,\ell}P_{d}
\;=\; \mathbf{X}^{\ell} P_{d},
\end{aligned}
\end{equation}

\emph{(6) Final normalization.} We have step (2) once more, with $\boldsymbol{\gamma}^{f}P_{d},
\boldsymbol{\beta}^{f}P_{d}$ and ${\mathbf{X}^{L}}' = \mathbf{X}^{L}P_{d}$, gives
${\tilde{\mathbf{X}}^{L}{}}' = \tilde{\mathbf{X}}^{L}P_{d}$.

\emph{(7) Unembedding.} The unembedding simply gives:
\begin{equation}
\label{eq:app-conclusion}
\mathbf{Z}' = (\tilde{\mathbf{X}}^{L} P_{d})(P_{d}^{\top} \mathbf{F} P_{V})
= \mathbf{Z}P_{V},
\end{equation}
hence:
\begin{equation}
f_{s \cdot w}\bigl(\pi_{V}(\mathbf{t})\bigr) = \softmax(\mathbf{Z})P_{V} = f_{w}(\mathbf{t})\,P_{V}.
\end{equation}


\subsection{Proof of \texorpdfstring{\Cref{thm:graph-isomorphism}}{Theorem 2}}
\label{app:graph-construction-proof}

\textbf{Proof overview.}
The proof is constructed as follows. We start by formally defining an isomorphism and an
automorphism of a directed attributed multigraph, the kind of graph produced by our constructions, and
the graph of a transformer without the weights on its edges, which is exactly the same for all
transformers.
Next, in (\Cref{lem:structure-gl}) we describe the set of all automorphisms of a transformer's graph in
which the model's weights are excluded from the edge attributes, i.e.\ every edge keeps its type but
not its weight. Intuitively, these automorphisms describe how elements such as coupling edges, node heads,
and node attributes, limit a bijection from being an automorphism of our graph construction.
Then, in (\Cref{lem:values-gl}) we describe how such an automorphism acts on the weight matrices sitting
on the edges of the graph: it is an isomorphism from the graph of one transformer to the graph of
another if and only if the weight matrices of the second are those of the first, with their rows and
columns permuted according to the automorphism.
Once all the above is established, we move on to prove
the theorem itself. Starting with two weight configurations of the same orbit, we build from the group
element relating them an automorphism of the weightless graph, and show by \Cref{lem:values-gl} that it is
an isomorphism between the two graphs they generate, proving these are isomorphic. Then,
given two isomorphic transformers' graphs, we first note that any isomorphism between them must also be
an automorphism of the transformer's graph without the weights on the edges, since two transformers'
graphs are entirely identical except for the weights on their edges. By \Cref{lem:structure-gl,lem:values-gl},
the weight matrices of the two transformers are then related by the node permutations of this
automorphism, from which we build a group element relating the two transformers, i.e.\ they belong to the
same orbit (for GMT-GL, the general-linear components of this element are recovered from the products of
the attention matrices, under a full-rank assumption on these matrices). We carry out this plan for GMT-GL (\Cref{app:proof-gl}), and
then describe the modifications each step requires for GMT-P (\Cref{app:proof-p}). Both constructions
share the preliminaries below.

\textbf{Directed attributed multigraphs.}
The graphs produced by our constructions are directed, carry attributes on both their nodes and their
edges, and may join the same pair of nodes by more than one edge (e.g.\ a weight edge and a coupling
edge), though there are no two edges with the same source, target, and attributes.
We therefore define isomorphisms and automorphisms for this kind of graph.
\begin{definition}[Isomorphism of directed attributed multigraphs]
\label{def:isomorphism}
A \emph{directed attributed multigraph} $G = (\mathcal{V}, \mathcal{E}, c)$ consists of a finite set of
nodes $\mathcal{V}$, a map $c$ assigning to every node $v \in \mathcal{V}$ a tuple of node attributes
$c(v)$, and a set of edges $\mathcal{E}$, where every edge is a triple $(u, v, X)$ of its source node
$u \in \mathcal{V}$, its target node $v \in \mathcal{V}$, and the tuple $X$ of its edge attributes;
several edges may join the same ordered pair of nodes, provided they carry different attributes. Let
$G = (\mathcal{V}, \mathcal{E}, c)$ and $G' = (\mathcal{V}', \mathcal{E}', c')$ be directed attributed
multigraphs. An \emph{isomorphism} from $G$ to $G'$ is a bijection of the nodes $F : \mathcal{V} \to
\mathcal{V}'$ such that
\begin{enumerate}
  \item for every node $v \in \mathcal{V}$, the nodes $F(v)$ and $v$ have exactly the same node
  attributes, $c'(F(v)) = c(v)$;
  \item for all nodes $u, v \in \mathcal{V}$ and every tuple of edge attributes $X$, the edge
  $(u, v, X)$ exists in $G$ if and only if the edge $(F(u), F(v), X)$ exists in $G'$.
\end{enumerate}
$G$ and $G'$ are \emph{isomorphic}, written $G \cong G'$, if there is an isomorphism from $G$ to $G'$.
\end{definition}
\begin{definition}[Automorphism of a directed attributed multigraph]
\label{def:automorphism}
Let $G = (\mathcal{V}, \mathcal{E}, c)$ be a directed attributed multigraph. An \emph{automorphism} of
$G$ is an isomorphism from $G$ to itself (\Cref{def:isomorphism}), i.e.\ a permutation of the nodes
$F : \mathcal{V} \to \mathcal{V}$ that preserves the node attributes and the edges of $G$, together
with their directions and attributes.
\end{definition}

\textbf{Conventions.}
Throughout the proof, the architecture, i.e.\ $L, H, d, d_k, d_v, d_{\mathrm{ff}}$ and $V$, is fixed,
as it is in the definition of $\mathcal{W}$ (\Cref{eq:weight-space}), and we use the notation of
\Cref{app:symmetries-proof}. Given $w \in \mathcal{W}$, we mark the parameters of a second weight
configuration $w' \in \mathcal{W}$ with a prime, e.g.\ $\mathbf{E}'$ or $(\mathbf{W}^{q,h,\ell})'$, and
write $\mathcal{G} \cdot w = \{ s \cdot w : s \in \mathcal{G} \}$ for the orbit of $w$. The permutation
matrix of $\pi \in S_n$ is $P_\pi$.

\textbf{Encodings.}
We write $\Gamma(w)$ for either $\Gamma_{\mathrm{GL}}(w)$ or $\Gamma_{\mathrm{P}}(w)$ when a statement
holds for both. Their nodes and edges carry the encodings of \Cref{sec:method-graph}:
\begin{enumerate}
  \item[(i)] Every node $v$ carries the encoding $c(v) = (\text{layer index}, \text{node type})$. All
  nodes of a node layer share their layer index and type, and different node layers have different
  layer indices, except that the $H$ per-head copies of a node layer within one block share their
  layer index and type. The scale node $g$ and the shift node $s$ of a normalization layer have
  different types.
  \item[(ii)] Every edge carries the encoding $X = (\text{weight}, \text{type})$. The edges of a
  fully-connected edge set $\mathcal{E}_{\crossarrow}(\mathcal{A}, \mathcal{B}; \mathbf{M})$ (\Cref{eq:fc-edges}),
  which we call \emph{weight edges}, share a type corresponding to the weight matrix they represent,
  and their weights are the corresponding entries of $\mathbf{M}$. This type does not depend on the
  head, so the corresponding edge sets of different heads in one block have the same type. The edges
  of a coupling edge set (\Cref{eq:coupling-edges}) have type \emph{coupling}, and the edges joining a
  head node to its head (GMT-GL only) have type \emph{head}; these parameter-free edges carry a fixed
  weight.
\end{enumerate}

\textbf{The graph without the weights.}
\begin{definition}[Weightless graph]
\label{def:weightless}
Given $w \in \mathcal{W}$, the \emph{weightless graph} of $\Gamma(w)$ is the graph obtained from
$\Gamma(w)$ by excluding the weights from the edge attributes: it has the same nodes and node encodings
as $\Gamma(w)$, and an edge $(u, v, \tau)$ for every edge $(u, v, (x, \tau))$ of $\Gamma(w)$. By the
graph construction (\Cref{sec:method-graph}), the nodes of $\Gamma(w)$, their encodings, and the
endpoints and types of its edges are determined by the architecture alone, and only the weights on its
edges depend on $w$. Hence, the weightless graphs of all $w \in \mathcal{W}$ are exactly the same graph,
an importantly, not up to a permutation of the nodes, i.e. they have exactly the same list of node encodings and exactly the same
adjacency matrix for every edge type.
From now on, we denote this weightless graph by $\bar\Gamma = (\mathcal{V}, \bar{\mathcal{E}}, c)$, and
by $\bar\Gamma_{\mathrm{GL}}$ or $\bar\Gamma_{\mathrm{P}}$ when referring to a specific construction.
Consequently, every graph $\Gamma(w)$ is the weightless graph $\bar\Gamma$
with a weight attached to each of its edges.
\end{definition}

\subsubsection{GMT-GL}
\label{app:proof-gl}

\textbf{The graph $\Gamma_{\mathrm{GL}}(w)$.}
We first characterize all nodes and
edges of this construction compactly in \Cref{tab:app-graph-gl}.

\begin{table}[H]
\caption{Node layers and edge sets of $\Gamma_{\mathrm{GL}}(w)$, for every block $\ell \in [L]$ and
head $h \in [H]$, grouped by the network component they represent and numbered as in
\Cref{tab:symmetries}. The residual layers $\mathcal{R}^{2\ell-2}, \mathcal{R}^{2\ell-1},
\mathcal{R}^{2\ell}$ are the layers $\mathcal{V}_d, \mathcal{V}_d', \mathcal{V}_d''$ of block $\ell$ in
\Cref{sec:GL-invariant-graph}, which we denote here differently to conveniently account for the nodes of different blocks. The nodes are named $\mathcal{V}_V = \{v_i\}$, $\mathcal{V}_V' =
\{v'_i\}$, $\mathcal{R}^k = \{r^k_i\}$, $\mathcal{V}_d^{h,\ell} = \{x^{h,\ell}_i\}$,
$\mathcal{V}_{ff}^\ell = \{z^\ell_i\}$ and $\mathcal{V}_{\gamma,\beta} = \{g, s\}$, with $u^{h,\ell}$
the head node of head $h$. The attention products are $\mathbf{M}^{h,\ell} = \mathbf{W}^{q,h,\ell}
(\mathbf{W}^{k,h,\ell})^{\top}$ and $\mathbf{N}^{h,\ell} = \mathbf{W}^{v,h,\ell}
\mathbf{W}^{O,\ell}_{h}$, and $[\boldsymbol{\gamma}; \boldsymbol{\beta}]$ is the $2 \times d$ matrix with
rows $\boldsymbol{\gamma}$ and $\boldsymbol{\beta}$. Fully-connected edge sets $\mathcal{E}_{\crossarrow}(\cdot, \cdot;
\cdot)$ consist of weight edges, coupling edge sets $\mathcal{E}_{\pararrow}(\cdot, \cdot)$ have type
\emph{coupling}, and the edges joining $u^{h,\ell}$ to its head have type \emph{head}.}
\label{tab:app-graph-gl}
\begin{center}
\small
\begin{tabular}{@{}llll@{}}
\toprule
\multicolumn{1}{c}{\bf Component} & \multicolumn{1}{c}{\bf Node layers} &
\multicolumn{1}{c}{\bf Weight edges} & \multicolumn{1}{c}{\bf Parameter-free edges} \\
\midrule
(1) Embedding & $\mathcal{V}_V$, $\mathcal{R}^{0}$
  & $\mathcal{E}_{\crossarrow}(\mathcal{V}_V, \mathcal{R}^{0}; \mathbf{E})$ & -- \\
\addlinespace[2pt]
(2) Pre-attn.\ norm & $\mathcal{V}_{\gamma,\beta}^{a,\ell}$
  & $\mathcal{E}_{\crossarrow}(\mathcal{V}_{\gamma,\beta}^{a,\ell}, \mathcal{R}^{2\ell-2};
    [\boldsymbol{\gamma}^{a,\ell}; \boldsymbol{\beta}^{a,\ell}])$ & -- \\
\addlinespace[2pt]
(3) Attention & $\mathcal{V}_d^{h,\ell}$, $u^{h,\ell}$, $\mathcal{R}^{2\ell-1}$
  & $\mathcal{E}_{\crossarrow}(\mathcal{R}^{2\ell-2}, \mathcal{V}_d^{h,\ell}; \mathbf{M}^{h,\ell})$
  & $\mathcal{E}_{\pararrow}(\mathcal{R}^{2\ell-2}, \mathcal{V}_d^{h,\ell})$ \\
 & & $\mathcal{E}_{\crossarrow}(\mathcal{V}_d^{h,\ell}, \mathcal{R}^{2\ell-1}; \mathbf{N}^{h,\ell})$
  & $\mathcal{E}_{\pararrow}(\mathcal{R}^{2\ell-2}, \mathcal{R}^{2\ell-1})$ \\
 & & & $\{(u^{h,\ell}, x^{h,\ell}_i) : i \in [d]\}$ \\
\addlinespace[2pt]
(4) Pre-MLP norm & $\mathcal{V}_{\gamma,\beta}^{b,\ell}$
  & $\mathcal{E}_{\crossarrow}(\mathcal{V}_{\gamma,\beta}^{b,\ell}, \mathcal{R}^{2\ell-1};
    [\boldsymbol{\gamma}^{b,\ell}; \boldsymbol{\beta}^{b,\ell}])$ & -- \\
\addlinespace[2pt]
(5) Feed-forward & $\mathcal{V}_{ff}^{\ell}$, $\mathcal{R}^{2\ell}$
  & $\mathcal{E}_{\crossarrow}(\mathcal{R}^{2\ell-1}, \mathcal{V}_{ff}^{\ell}; \mathbf{W}^{1,\ell})$
  & $\mathcal{E}_{\pararrow}(\mathcal{R}^{2\ell-1}, \mathcal{R}^{2\ell})$ \\
 & & $\mathcal{E}_{\crossarrow}(\mathcal{V}_{ff}^{\ell}, \mathcal{R}^{2\ell}; \mathbf{W}^{2,\ell})$ & \\
\addlinespace[2pt]
(6) Final norm & $\mathcal{V}_{\gamma,\beta}^{f}$
  & $\mathcal{E}_{\crossarrow}(\mathcal{V}_{\gamma,\beta}^{f}, \mathcal{R}^{2L};
    [\boldsymbol{\gamma}^{f}; \boldsymbol{\beta}^{f}])$ & -- \\
\addlinespace[2pt]
(7) Unembedding & $\mathcal{V}_V'$
  & $\mathcal{E}_{\crossarrow}(\mathcal{R}^{2L}, \mathcal{V}_V'; \mathbf{F})$
  & $\mathcal{E}_{\pararrow}(\mathcal{V}_V, \mathcal{V}_V')$ \\
\bottomrule
\end{tabular}
\end{center}
\end{table}

\textbf{Automorphisms of the graph without the weights.}
We first describe the set of all automorphisms of the weightless graph $\bar\Gamma_{\mathrm{GL}}$. By
\Cref{def:isomorphism,def:automorphism}, these are the bijections $F : \mathcal{V} \to \mathcal{V}$ that preserve the
node encodings and the typed edges: for every edge type $\tau$, there is an edge of type $\tau$ from
$u$ to $v$ if and only if there is an edge of type $\tau$ from $F(u)$ to $F(v)$.

\begin{lemma}
\label{lem:structure-gl}
A bijection $F : \mathcal{V} \to \mathcal{V}$ is an automorphism of the weightless graph
$\bar\Gamma_{\mathrm{GL}}$ if and only if there are permutations $\pi_V \in S_V$, $\pi_d \in S_d$ and,
for every block $\ell \in [L]$, $\rho_\ell \in S_H$ and $\pi^{\ell}_{\mathrm{ff}} \in
S_{d_{\mathrm{ff}}}$, such that $F$ fixes every normalization node and, for all blocks $\ell$, heads
$h$, $k \in \{0, \dots, 2L\}$ and indices $i$,
\begin{equation}
\label{eq:app-structure-gl}
\begin{aligned}
F(v_i) &= v_{\pi_V(i)}, & F(v'_i) &= v'_{\pi_V(i)}, & F(r^k_i) &= r^k_{\pi_d(i)}, \\
F(u^{h,\ell}) &= u^{\rho_\ell(h),\ell}, & F(x^{h,\ell}_i) &= x^{\rho_\ell(h),\ell}_{\pi_d(i)}, &
F(z^{\ell}_i) &= z^{\ell}_{\pi^{\ell}_{\mathrm{ff}}(i)} .
\end{aligned}
\end{equation}
We denote this map by $F^{\mathrm{GL}}_\theta$, where $\theta = \bigl(\pi_V, \pi_d, (\rho_\ell,
\pi^{\ell}_{\mathrm{ff}})_{\ell \in [L]}\bigr)$.
\end{lemma}
That is, an automorphism of the weightless graph consists of one vocabulary permutation shared by both
embedding and unembedding layers, one feature permutation shared by all residual and per-head layers,
and, in every block, a permutation of the heads and a permutation of the hidden neurons.

\begin{proof}
We first show that every automorphism of $\bar\Gamma_{\mathrm{GL}}$ has this form, and then that every
map of this form is an automorphism of $\bar\Gamma_{\mathrm{GL}}$. Let $F$ be an automorphism of
$\bar\Gamma_{\mathrm{GL}}$; we go over the graph layer by layer, following the components of
\Cref{tab:app-graph-gl}.

Before investigation every layer, describe how node types limits the automorphism and how different connectivity patterns constrain the action of an automorphism on the nodes. 

\emph{Encodings.} Since $F$ preserves the node encodings, it maps every set of nodes sharing one
encoding into itself, and hence onto itself, as $F$ is injective and the set is finite. By (i), these
sets are $\mathcal{V}_V$, $\mathcal{V}_V'$, every $\mathcal{R}^k$, every $\mathcal{V}_{ff}^\ell$, every
normalization node by itself, and, for every block $\ell$, the union $\bigcup_h \mathcal{V}_d^{h,\ell}$
of its per-head layers and the set $\{u^{1,\ell}, \dots, u^{H,\ell}\}$ of its head nodes. When $F$ maps
a node layer $\mathcal{A} = \{a_i\}$ onto a node layer $\mathcal{A}' = \{a'_i\}$, we call the
permutation $\alpha$ with $F(a_i) = a'_{\alpha(i)}$ the \emph{index permutation} of $F$ on
$\mathcal{A}$. Fully-connected edge sets join every node of one layer to every node of another, so they
impose no constraint beyond the encodings; the constraints come from the coupling and head edges.

\emph{Coupled layers share their index permutation.} Let $\mathcal{E}_{\pararrow}(\mathcal{A}, \mathcal{B})$
be a coupling edge set, and suppose that $F$ maps $\mathcal{A}$ and $\mathcal{B}$ onto node layers
$\mathcal{A}'$ and $\mathcal{B}'$ that are joined by the coupling edge set $\mathcal{E}_{\pararrow}(\mathcal{A}',
\mathcal{B}')$, with index permutations $\alpha$ and $\beta$. The coupling edge $(a_i, b_i)$ is mapped
to $(a'_{\alpha(i)}, b'_{\beta(i)})$, which must be a coupling edge as well. Since the only coupling
edges from $\mathcal{A}'$ to $\mathcal{B}'$ are those of $\mathcal{E}_{\pararrow}(\mathcal{A}', \mathcal{B}')$,
which join nodes of equal index, $\alpha(i) = \beta(i)$ for all $i$, i.e.\ $\alpha = \beta$.

We now determine $F$ on every component of the graph.

\emph{(1) Embedding.} By the encodings, $F$ maps $\mathcal{V}_V$ and $\mathcal{R}^0$ onto themselves;
let $\pi_V$ and $\pi_d$ be its index permutations on these node layers.
Since there is only a fully connected edge set joining $\mathcal{V}_V$ to $\mathcal{R}^0$,
there are no constraints on $\pi_V$ and $\pi_d$ beyond the encodings.
Later on, we show that these two permutations are shared by the unembedding layer and all residual and per-head layers.

\emph{(2) Pre-attention normalization.} Each of $g^{a,\ell}$ and $s^{a,\ell}$ is the only node with its
encoding, so $F$ fixes both.

We treat the blocks in order, $\ell = 1, \dots, L$, assuming that the index permutation of $F$ on the
input layer $\mathcal{R}^{2\ell-2}$ of block $\ell$ is $\pi_d$ and prove it via induction. For $\ell = 1$, this is the definition
of $\pi_d$, then for any $\ell \ge 1$  we assume it is true, and later prove it also holds for $\ell +1$.

\emph{(3) Attention.} By the encodings, $F$ permutes the head nodes of block $\ell$, $F(u^{h,\ell}) =
u^{\rho_\ell(h),\ell}$ for some $\rho_\ell \in S_H$. The head node $u^{h,\ell}$ is connected to 
the nodes of $\mathcal{V}_d^{h,\ell}$ and only to them. Hence, for every $i$, the head edge $(u^{h,\ell},
x^{h,\ell}_i)$ is mapped to $(u^{\rho_\ell(h),\ell}, F(x^{h,\ell}_i))$, which must be a head edge as
well, so $F(x^{h,\ell}_i) \in \mathcal{V}_d^{\rho_\ell(h),\ell}$. Thus $F$ maps $\mathcal{V}_d^{h,\ell}$
 onto, $\mathcal{V}_d^{\rho_\ell(h),\ell}$; that is, the head node forces the $d$ nodes of its head
to move together to a single head, rather than each of them being moved to a head of its own. Next,
$\mathcal{V}_d^{h,\ell}$ is coupled to $\mathcal{R}^{2\ell-2}$, and $F$ maps these two layers onto
$\mathcal{V}_d^{\rho_\ell(h),\ell}$ and $\mathcal{R}^{2\ell-2}$, which are coupled as well. Hence the
index permutation of $F$ on $\mathcal{V}_d^{h,\ell}$ is that on $\mathcal{R}^{2\ell-2}$, namely $\pi_d$,
i.e.\ $F(x^{h,\ell}_i) = x^{\rho_\ell(h),\ell}_{\pi_d(i)}$. Finally, the attention output
$\mathcal{R}^{2\ell-1}$ is coupled to $\mathcal{R}^{2\ell-2}$ by the residual connection, so its index
permutation is $\pi_d$ as well.

\emph{(4) Pre-MLP normalization.} As in (2), $g^{b,\ell}$ and $s^{b,\ell}$ are fixed.

\emph{(5) Feed-forward.} Let $\pi^\ell_{\mathrm{ff}}$ be the index permutation of $F$ on
$\mathcal{V}^\ell_{ff}$; only fully-connected edge sets are incident to $\mathcal{V}^\ell_{ff}$, so it is
not constrained further. The layer $\mathcal{R}^{2\ell}$ is coupled to $\mathcal{R}^{2\ell-1}$ by the
residual connection, so its index permutation is $\pi_d$. As $\mathcal{R}^{2\ell}$ is the input layer of
block $\ell + 1$, this completes the induction that states that for each $\ell$, .

\emph{(6) Final normalization.} As in (2), $g^f$ and $s^f$ are fixed.

\emph{(7) Unembedding.} The layer $\mathcal{V}_V'$ is coupled to $\mathcal{V}_V$, so its index
permutation is $\pi_V$.

We have now determined $F$ on every node of the graph, and it acts as in \Cref{eq:app-structure-gl},
i.e.\ $F = F^{\mathrm{GL}}_\theta$.

Conversely, let $F^{\mathrm{GL}}_\theta$ be as in \Cref{eq:app-structure-gl}.
It clearly maps every node to a node with the
same encoding (which is constructed of type and layer), which is the first condition of \Cref{def:isomorphism}.
Moving on to the edges, we first note that $F^{\mathrm{GL}}_\theta$ maps every node layer to a node layer of the same layer and type,
then since a weight edge is part of a fully connected edge set, clearly the mapped edge also exists and of the same type.
Next, for coupling edges, we have already shown that the index permutation of $F^{\mathrm{GL}}_\theta$ is the same on both layers joined by a coupling edge set,
hence the mapped edge also exists and of the same type.
Finally, for head edges, the head node of a head is mapped to the head node of another head, together with all nodes associated with that head, so again the mapped edge also exists and of the same type.
After considering all nodes and edges, we see that by definition $F^{\mathrm{GL}}_\theta$ is indeed an automorphism of $\bar\Gamma_{\mathrm{GL}}$.
\end{proof}

\textbf{The action of an automorphism on the weights.}
Next, we describe how an automorphism $F^{\mathrm{GL}}_\theta$ of the weightless graph 
acts on the weight matrices, and consequently, under what conditions it is an isomorphism between the graphs of two transformers.
sitting on the edges of the graph.

From now, we sometimes use the shorthand
$\mathbf{M}^{h,\ell} = \mathbf{W}^{q,h,\ell} (\mathbf{W}^{k,h,\ell})^{\top}$ and
$\mathbf{N}^{h,\ell} = \mathbf{W}^{v,h,\ell} \mathbf{W}^{O,\ell}_{h}$ for the query--key and
value--output products of head $h$ in block $\ell$ of $w$.

\begin{lemma}
\label{lem:values-gl}
Let $w, w' \in \mathcal{W}$, and let $F^{\mathrm{GL}}_\theta$ be as in \Cref{lem:structure-gl}. Then $F^{\mathrm{GL}}_\theta$ is an
isomorphism from $\Gamma_{\mathrm{GL}}(w)$ to $\Gamma_{\mathrm{GL}}(w')$ if and only if
\begin{equation}
\label{eq:app-iso-gl}
\begin{gathered}
\mathbf{E}' = P_V^{\top} \mathbf{E} P_d,
\qquad
\mathbf{F}' = P_d^{\top} \mathbf{F} P_V,
\qquad
[\boldsymbol{\gamma}'; \boldsymbol{\beta}'] = [\boldsymbol{\gamma}; \boldsymbol{\beta}]\, P_d
\ \text{ for every normalization layer},
\\
(\mathbf{W}^{1,\ell})' = P_d^{\top} \mathbf{W}^{1,\ell} \hat{P}^{\ell},
\qquad
(\mathbf{W}^{2,\ell})' = (\hat{P}^{\ell})^{\top} \mathbf{W}^{2,\ell} P_d,
\\
(\mathbf{M}^{h,\ell})' = P_d^{\top} \mathbf{M}^{\sigma_\ell(h),\ell} P_d,
\qquad
(\mathbf{N}^{h,\ell})' = P_d^{\top} \mathbf{N}^{\sigma_\ell(h),\ell} P_d,
\end{gathered}
\end{equation}
for every block $\ell \in [L]$ and head $h \in [H]$.
\end{lemma}

\begin{proof}
By \Cref{lem:structure-gl}, $F^{\mathrm{GL}}_\theta$ is an automorphism of $\bar\Gamma_{\mathrm{GL}}$, which by
\Cref{def:weightless} is the weightless graph of both $\Gamma_{\mathrm{GL}}(w)$ and
$\Gamma_{\mathrm{GL}}(w')$. Hence $F^{\mathrm{GL}}_\theta$ maps every node to a node with the same encoding, which is
the first condition of \Cref{def:isomorphism}, and it maps the edges of $\Gamma_{\mathrm{GL}}(w)$
one-to-one onto the edges of $\Gamma_{\mathrm{GL}}(w')$, each to an edge of the same type. The second
condition of \Cref{def:isomorphism} therefore holds if and only if every edge of
$\Gamma_{\mathrm{GL}}(w)$ is mapped to an edge of $\Gamma_{\mathrm{GL}}(w')$ that carries the same
weight. The parameter-free edges carry the same fixed weight in both graphs, so it remains to consider
the weight edges. Consider a fully-connected edge set $\mathcal{E}_{\crossarrow}(\mathcal{A}, \mathcal{B};
\mathbf{M})$ of $\Gamma_{\mathrm{GL}}(w)$, let $F^{\mathrm{GL}}_\theta$ map $\mathcal{A} = \{a_i\}$ onto
$\mathcal{A}' = \{a'_i\}$ and $\mathcal{B} = \{b_j\}$ onto $\mathcal{B}' = \{b'_j\}$, with index
permutations $\alpha$ and $\beta$, and let $\mathbf{M}'$ be the matrix carried by the edge set from
$\mathcal{A}'$ to $\mathcal{B}'$ in $\Gamma_{\mathrm{GL}}(w')$. The edge $(a_i, b_j)$, which carries the
weight $\mathbf{M}_{i,j}$, is mapped to the edge $(a'_{\alpha(i)}, b'_{\beta(j)})$, which carries the
weight $\mathbf{M}'_{\alpha(i), \beta(j)}$. Hence the weights on this edge set are preserved if and only
if $\mathbf{M}'_{\alpha(i), \beta(j)} = \mathbf{M}_{i,j}$ for all $i, j$, that is:
\begin{equation}
\label{eq:app-move-matrix}
\mathbf{M}' = P_\alpha^{\top} \mathbf{M} P_\beta .
\end{equation}
We apply \Cref{eq:app-move-matrix} to every weight edge set of \Cref{tab:app-graph-gl}, reading the
index permutations off \Cref{eq:app-structure-gl}:
\begin{itemize}
  \item[(1)] \emph{Embedding.} $\mathbf{E}$ sits between $\mathcal{V}_V$ ($\pi_V$) and $\mathcal{R}^0$
  ($\pi_d$), giving $\mathbf{E}' = P_V^{\top} \mathbf{E} P_d$. Concretely, the edge $(v_i, r^0_j)$,
  which carries $\mathbf{E}_{i,j}$, is mapped to the edge $(v_{\pi_V(i)}, r^0_{\pi_d(j)})$, which carries
  $\mathbf{E}'_{\pi_V(i), \pi_d(j)}$.
  \item[(2)] \emph{Pre-attention normalization.} $[\boldsymbol{\gamma}^{a,\ell};
  \boldsymbol{\beta}^{a,\ell}]$ sits between $\mathcal{V}^{a,\ell}_{\gamma,\beta}$, whose two nodes are
  fixed (permutation matrix $I_2$), and $\mathcal{R}^{2\ell-2}$ ($\pi_d$), giving
  $[(\boldsymbol{\gamma}^{a,\ell})'; (\boldsymbol{\beta}^{a,\ell})'] = [\boldsymbol{\gamma}^{a,\ell};
  \boldsymbol{\beta}^{a,\ell}]\, P_d$.
  \item[(3)] \emph{Attention.} $\mathbf{M}^{j,\ell}$ sits between $\mathcal{R}^{2\ell-2}$ ($\pi_d$) and
  $\mathcal{V}_d^{j,\ell}$, which $F^{\mathrm{GL}}_\theta$ maps onto $\mathcal{V}_d^{\rho_\ell(j),\ell}$ with index
  permutation $\pi_d$. In $\Gamma_{\mathrm{GL}}(w')$, the edge set from $\mathcal{R}^{2\ell-2}$ to
  $\mathcal{V}_d^{\rho_\ell(j),\ell}$ carries $(\mathbf{M}^{\rho_\ell(j),\ell})'$, giving
  $(\mathbf{M}^{\rho_\ell(j),\ell})' = P_d^{\top} \mathbf{M}^{j,\ell} P_d$, i.e.\ $(\mathbf{M}^{h,\ell})' =
  P_d^{\top} \mathbf{M}^{\sigma_\ell(h),\ell} P_d$ for $h = \rho_\ell(j)$. Likewise,
  $\mathbf{N}^{j,\ell}$ sits between $\mathcal{V}_d^{j,\ell}$ (mapped onto
  $\mathcal{V}_d^{\rho_\ell(j),\ell}$, $\pi_d$) and $\mathcal{R}^{2\ell-1}$ ($\pi_d$), giving
  $(\mathbf{N}^{h,\ell})' = P_d^{\top} \mathbf{N}^{\sigma_\ell(h),\ell} P_d$.
  \item[(4)] \emph{Pre-MLP normalization.} As in (2), with $\mathcal{R}^{2\ell-1}$ ($\pi_d$):
  $[(\boldsymbol{\gamma}^{b,\ell})'; (\boldsymbol{\beta}^{b,\ell})'] = [\boldsymbol{\gamma}^{b,\ell};
  \boldsymbol{\beta}^{b,\ell}]\, P_d$.
  \item[(5)] \emph{Feed-forward.} $\mathbf{W}^{1,\ell}$ sits between $\mathcal{R}^{2\ell-1}$ ($\pi_d$)
  and $\mathcal{V}^\ell_{ff}$ ($\pi^\ell_{\mathrm{ff}}$), and $\mathbf{W}^{2,\ell}$ between
  $\mathcal{V}^\ell_{ff}$ ($\pi^\ell_{\mathrm{ff}}$) and $\mathcal{R}^{2\ell}$ ($\pi_d$), giving
  $(\mathbf{W}^{1,\ell})' = P_d^{\top} \mathbf{W}^{1,\ell} \hat{P}^{\ell}$ and $(\mathbf{W}^{2,\ell})' =
  (\hat{P}^{\ell})^{\top} \mathbf{W}^{2,\ell} P_d$.
  \item[(6)] \emph{Final normalization.} As in (2), with $\mathcal{R}^{2L}$ ($\pi_d$):
  $[(\boldsymbol{\gamma}^{f})'; (\boldsymbol{\beta}^{f})'] = [\boldsymbol{\gamma}^{f};
  \boldsymbol{\beta}^{f}]\, P_d$.
  \item[(7)] \emph{Unembedding.} $\mathbf{F}$ sits between $\mathcal{R}^{2L}$ ($\pi_d$) and
  $\mathcal{V}_V'$ ($\pi_V$), giving $\mathbf{F}' = P_d^{\top} \mathbf{F} P_V$.
\end{itemize}
These are exactly the equations of \Cref{eq:app-iso-gl}, so $F^{\mathrm{GL}}_\theta$ is an isomorphism from
$\Gamma_{\mathrm{GL}}(w)$ to $\Gamma_{\mathrm{GL}}(w')$ if and only if they hold.
\end{proof}

\textbf{Proof of \Cref{thm:graph-isomorphism} for GMT-GL.}
We now prove the two directions of the theorem for GMT-GL.

\emph{First direction: transformers from the same orbit generate isomorphic graphs.}
Let $w \in \mathcal{W}$ and $s \in \mathcal{G}$, and write $w' = s \cdot w$; we build an isomorphism from
$\Gamma_{\mathrm{GL}}(w)$ to $\Gamma_{\mathrm{GL}}(w')$. Let $P_V$, $P_d$, $\sigma_\ell$, $\hat{P}^\ell$
and $G^{h,\ell}_{qk}, G^{h,\ell}_{vo}$ be the components of $s$ (\Cref{tab:symmetries}), and let
$\pi_V$, $\pi_d$ and $\pi^\ell_{\mathrm{ff}}$ be permutations such that $P_V = P_{\pi_V}$, $P_d =
P_{\pi_d}$ and $\hat{P}^\ell = P_{\pi^\ell_{\mathrm{ff}}}$. We define
\[
\theta = \bigl(\pi_V, \pi_d, (\sigma_\ell^{-1}, \pi^\ell_{\mathrm{ff}})_{\ell \in [L]}\bigr),
\]
i.e.\ $\rho_\ell = \sigma_\ell^{-1}$ in \Cref{lem:structure-gl}, so that the matrices $P_V$, $P_d$,
$\hat{P}^\ell$ and the head permutations $\sigma_\ell$ associated with $\theta$ in
\Cref{lem:values-gl} are exactly the permutation components of $s$. By \Cref{lem:structure-gl},
$F^{\mathrm{GL}}_\theta$ is an automorphism of $\bar\Gamma_{\mathrm{GL}}$, and we now verify that $w$ and $w' = s
\cdot w$ satisfy \Cref{eq:app-iso-gl}. By rows (1), (2) and (4)--(7) of \Cref{tab:symmetries},
$\mathbf{E}' = P_V^{\top} \mathbf{E} P_d$, $\mathbf{F}' = P_d^{\top} \mathbf{F} P_V$,
$[\boldsymbol{\gamma}'; \boldsymbol{\beta}'] = [\boldsymbol{\gamma}; \boldsymbol{\beta}]\, P_d$ for
every normalization layer, $(\mathbf{W}^{1,\ell})' = P_d^{\top} \mathbf{W}^{1,\ell} \hat{P}^{\ell}$ and
$(\mathbf{W}^{2,\ell})' = (\hat{P}^{\ell})^{\top} \mathbf{W}^{2,\ell} P_d$, which are the first two lines
of \Cref{eq:app-iso-gl}. For the attention, fix $\ell$ and $h$, let $j = \sigma_\ell(h)$, and
abbreviate $G_{qk} = G^{j,\ell}_{qk}$ and $G_{vo} = G^{j,\ell}_{vo}$. By row (3) of
\Cref{tab:symmetries}, the products of head $h$ in block $\ell$ of $w'$ are
\begin{align*}
(\mathbf{M}^{h,\ell})' = (\mathbf{W}^{q,h,\ell})' (\mathbf{W}^{k,h,\ell})'^{\top}
&= \bigl(P_d^{\top} \mathbf{W}^{q,j,\ell} G_{qk}\bigr)
  \bigl(G_{qk}^{-1} (\mathbf{W}^{k,j,\ell})^{\top} P_d\bigr)
= P_d^{\top} \mathbf{M}^{j,\ell} P_d,
\\
(\mathbf{N}^{h,\ell})' = (\mathbf{W}^{v,h,\ell})' (\mathbf{W}^{O,\ell}_{h})'
&= \bigl(P_d^{\top} \mathbf{W}^{v,j,\ell} G_{vo}\bigr)
  \bigl(G_{vo}^{-1} \mathbf{W}^{O,\ell}_{j} P_d\bigr)
= P_d^{\top} \mathbf{N}^{j,\ell} P_d,
\end{align*}
since the general-linear factors cancel; this is the last line of \Cref{eq:app-iso-gl}. Hence, by
\Cref{lem:values-gl}, $F^{\mathrm{GL}}_\theta$ is an isomorphism from $\Gamma_{\mathrm{GL}}(w)$ to
$\Gamma_{\mathrm{GL}}(w')$, and $\Gamma_{\mathrm{GL}}(w) \cong \Gamma_{\mathrm{GL}}(s \cdot w)$. Note that
the general-linear components of $s$ play no role in $F^{\mathrm{GL}}_\theta$: they cancel in the products of the
attention matrices, which is all that the graph carries.

\emph{Recovering the attention matrices from their products.}
For the second direction, we need to account for the general-linear symmetry, and recover the
individual attention matrices from their products, up to this symmetry. This is possible under the
following full-rank assumption on $w$, which makes the assumption of \Cref{thm:graph-isomorphism}
precise:
\begin{equation}
\label{eq:app-rank}
\operatorname{rank} \mathbf{W}^{q,h,\ell} = \operatorname{rank} \mathbf{W}^{k,h,\ell} = d_k,
\qquad
\operatorname{rank} \mathbf{W}^{v,h,\ell} = \operatorname{rank} \mathbf{W}^{O,\ell}_{h} = d_v,
\qquad \ell \in [L],\ h \in [H],
\end{equation}
i.e.\ $\mathbf{W}^{q,h,\ell}, \mathbf{W}^{k,h,\ell}, \mathbf{W}^{v,h,\ell}$ have full column rank and
$\mathbf{W}^{O,\ell}_h$ has full row rank. This requires $d_k, d_v \le d$, as in standard
architectures.

\begin{lemma}
\label{lem:factorization}
Let $\mathbf{A}, \mathbf{B}, \mathbf{A}', \mathbf{B}' \in \mathbb{R}^{d \times r}$ have rank $r$, and
satisfy $\mathbf{A}' \mathbf{B}'^{\top} = \mathbf{A} \mathbf{B}^{\top}$. Then there is $G \in
\mathrm{GL}(r)$ such that $\mathbf{A}' = \mathbf{A} G$ and $\mathbf{B}' = \mathbf{B} G^{-\top}$.
\end{lemma}
\begin{proof}
Since $\mathbf{B}'$ has rank $r$, the $r \times r$ matrix $\mathbf{B}'^{\top} \mathbf{B}'$ is invertible.
Multiplying $\mathbf{A}' \mathbf{B}'^{\top} = \mathbf{A} \mathbf{B}^{\top}$ on the right by
$\mathbf{B}' (\mathbf{B}'^{\top} \mathbf{B}')^{-1}$ gives
\[
\mathbf{A}' = \mathbf{A} G,
\qquad
G = \mathbf{B}^{\top} \mathbf{B}' (\mathbf{B}'^{\top} \mathbf{B}')^{-1} \in \mathbb{R}^{r \times r},
\]
and $G$ is invertible, since $r = \operatorname{rank} \mathbf{A}' = \operatorname{rank} (\mathbf{A} G)
\le \operatorname{rank} G$. Substituting $\mathbf{A}' = \mathbf{A} G$ back gives $\mathbf{A} G
\mathbf{B}'^{\top} = \mathbf{A} \mathbf{B}^{\top}$. Since $\mathbf{A}$ has rank $r$, the matrix
$\mathbf{A}^{\top} \mathbf{A}$ is invertible, and multiplying on the left by $(\mathbf{A}^{\top}
\mathbf{A})^{-1} \mathbf{A}^{\top}$ gives $G \mathbf{B}'^{\top} = \mathbf{B}^{\top}$, i.e.\ $\mathbf{B}' =
\mathbf{B} G^{-\top}$.
\end{proof}

\emph{Second direction: isomorphic graphs are generated by transformers from the same orbit.}
Let $w, w' \in \mathcal{W}$, where $w$ satisfies \Cref{eq:app-rank} and $\Gamma_{\mathrm{GL}}(w) \cong
\Gamma_{\mathrm{GL}}(w')$; we prove that $w' \in \mathcal{G} \cdot w$. Let $F$ be an isomorphism from
$\Gamma_{\mathrm{GL}}(w)$ to $\Gamma_{\mathrm{GL}}(w')$.

First, by
\Cref{def:weightless}, the weightless graphs of $\Gamma_{\mathrm{GL}}(w)$ and $\Gamma_{\mathrm{GL}}(w')$
are exactly the same graph $\bar\Gamma_{\mathrm{GL}} = (\mathcal{V}, \bar{\mathcal{E}}, c)$, and not
merely up to a permutation of the nodes. 
Therefore, it must be that $F$ is an automorphism of the weightless graph $\bar\Gamma_{\mathrm{GL}}$,
since $\Gamma_{\mathrm{GL}}(w), \Gamma_{\mathrm{GL}}(w')$ are just the graph $\bar\Gamma_{\mathrm{GL}}$ with additional weight attribute over the edges.

Now, by \Cref{lem:structure-gl}, $F = F^{\mathrm{GL}}_\theta$ for some $\theta = (\pi_V, \pi_d, (\rho_\ell,
\pi^\ell_{\mathrm{ff}})_{\ell \in [L]})$, and by \Cref{lem:values-gl}, $w$ and $w'$ satisfy
\Cref{eq:app-iso-gl}, with $P_V = P_{\pi_V}$, $P_d = P_{\pi_d}$, $\hat{P}^\ell =
P_{\pi^\ell_{\mathrm{ff}}}$ and $\sigma_\ell = \rho_\ell^{-1}$. It remains to build from $F^{\mathrm{GL}}_\theta$ a
group element $s \in \mathcal{G}$ with $w' = s \cdot w$. We take the permutation components of $s$ to be
$P_V$, $P_d$ and, for every block, $\sigma_\ell$ and $\hat{P}^\ell$. Then the first two lines of
\Cref{eq:app-iso-gl} are rows (1), (2) and (4)--(7) of \Cref{tab:symmetries}, and it remains to find
general-linear components of $s$ for which row (3) holds; these are recovered from the last line of
\Cref{eq:app-iso-gl} using \Cref{lem:factorization}.

Fix $\ell$ and $h$, and let $j = \sigma_\ell(h)$. For the query--key pair, apply
\Cref{lem:factorization} with $r = d_k$ and
\[
\mathbf{A} = P_d^{\top} \mathbf{W}^{q,j,\ell},
\quad
\mathbf{B} = P_d^{\top} \mathbf{W}^{k,j,\ell},
\quad
\mathbf{A}' = (\mathbf{W}^{q,h,\ell})',
\quad
\mathbf{B}' = (\mathbf{W}^{k,h,\ell})'.
\]
Here $\mathbf{A}$ and $\mathbf{B}$ have rank $d_k$ by \Cref{eq:app-rank}, as $P_d^{\top}$ is
invertible, and
\[
\mathbf{A}' \mathbf{B}'^{\top} = (\mathbf{M}^{h,\ell})' = P_d^{\top} \mathbf{M}^{j,\ell}
P_d = \mathbf{A} \mathbf{B}^{\top}.
\]
The latter product has rank $d_k$, and since
$\operatorname{rank}(\mathbf{A}' \mathbf{B}'^{\top}) \le \min(\operatorname{rank} \mathbf{A}',
\operatorname{rank} \mathbf{B}') \le d_k$, also $\mathbf{A}'$ and $\mathbf{B}'$ have rank $d_k$; that
is, the full-rank assumption on $w$ carries over to $w'$. The lemma yields $G^{j,\ell}_{qk} \in \mathrm{GL}(d_k)$ with
\[
(\mathbf{W}^{q,h,\ell})' = P_d^{\top} \mathbf{W}^{q,j,\ell} G^{j,\ell}_{qk},
\qquad
(\mathbf{W}^{k,h,\ell})' = P_d^{\top} \mathbf{W}^{k,j,\ell} \bigl(G^{j,\ell}_{qk}\bigr)^{-\top}.
\]
For the value--output pair, apply \Cref{lem:factorization} with $r = d_v$ and $\mathbf{A} = P_d^{\top}
\mathbf{W}^{v,j,\ell}$, $\mathbf{B} = P_d^{\top} (\mathbf{W}^{O,\ell}_{j})^{\top}$, $\mathbf{A}' =
(\mathbf{W}^{v,h,\ell})'$ and $\mathbf{B}' = (\mathbf{W}^{O,\ell}_{h})'^{\top}$, for which again
$\mathbf{A}' \mathbf{B}'^{\top} = (\mathbf{N}^{h,\ell})' = P_d^{\top} \mathbf{N}^{j,\ell} P_d =
\mathbf{A} \mathbf{B}^{\top}$, and all four matrices have rank $d_v$ by the same argument. It yields $G^{j,\ell}_{vo} \in \mathrm{GL}(d_v)$ with
\[
(\mathbf{W}^{v,h,\ell})' = P_d^{\top} \mathbf{W}^{v,j,\ell} G^{j,\ell}_{vo},
\qquad
(\mathbf{W}^{O,\ell}_{h})' = \bigl(G^{j,\ell}_{vo}\bigr)^{-1} \mathbf{W}^{O,\ell}_{j} P_d .
\]
As $h$ ranges over $[H]$, so does $j = \sigma_\ell(h)$, exactly once; hence every $G^{j,\ell}_{qk}$
and $G^{j,\ell}_{vo}$ is defined exactly once. These equations are row (3) of \Cref{tab:symmetries}.
Hence $w' = s \cdot w$ for the element $s \in \mathcal{G}$ with components $P_V$, $P_d$ and, for every
block, $\sigma_\ell$, $(G^{j,\ell}_{qk}, G^{j,\ell}_{vo})_{j \in [H]}$ and $\hat{P}^\ell$, i.e.\ $w$ and
$w'$ belong to the same orbit. This completes the proof of the GMT-GL part of
\Cref{thm:graph-isomorphism}.

\subsubsection{GMT-P}
\label{app:proof-p}

We follow the same plan used with GMT-GL, and describe the modifications each step requires.

\textbf{The graph $\Gamma_{\mathrm{P}}(w)$.}
The graph $\Gamma_{\mathrm{P}}(w)$ coincides with $\Gamma_{\mathrm{GL}}(w)$ in components (1), (2) and
(4)--(7) of \Cref{tab:app-graph-gl}, i.e.\ outside the attention sub-layers. In the attention of block
$\ell$ there are no head nodes and no product matrices. Instead, every head $h$ has three node layers,
$\mathcal{V}_{d_k}^{h,\ell} = \{y^{h,\ell}_i\}_{i=1}^{d_k}$, $\mathcal{V}_{d}^{h,\ell} =
\{x^{h,\ell}_i\}_{i=1}^{d}$ and $\mathcal{V}_{d_v}^{h,\ell} = \{w^{h,\ell}_i\}_{i=1}^{d_v}$, and every
attention matrix sits on its own weight edge set:
\[
\mathcal{R}^{2\ell-2} \xrightarrow{\;\mathbf{W}^{q,h,\ell}\;} \mathcal{V}_{d_k}^{h,\ell}
\xrightarrow{\;(\mathbf{W}^{k,h,\ell})^{\top}\;} \mathcal{V}_{d}^{h,\ell}
\xrightarrow{\;\mathbf{W}^{v,h,\ell}\;} \mathcal{V}_{d_v}^{h,\ell}
\xrightarrow{\;\mathbf{W}^{O,\ell}_{h}\;} \mathcal{R}^{2\ell-1},
\]
where an arrow $\mathcal{A} \xrightarrow{\mathbf{M}} \mathcal{B}$ denotes the fully-connected edge set
$\mathcal{E}_{\crossarrow}(\mathcal{A}, \mathcal{B}; \mathbf{M})$. The coupling edge sets
$\mathcal{E}_{\pararrow}(\mathcal{R}^{2\ell-2}, \mathcal{V}_d^{h,\ell})$ and
$\mathcal{E}_{\pararrow}(\mathcal{R}^{2\ell-2}, \mathcal{R}^{2\ell-1})$ are as in GMT-GL. By (i), the encoding
classes inside the attention of block $\ell$ are $\bigcup_h \mathcal{V}_{d_k}^{h,\ell}$, $\bigcup_h
\mathcal{V}_{d}^{h,\ell}$ and $\bigcup_h \mathcal{V}_{d_v}^{h,\ell}$; all other encoding classes are
as in GMT-GL. As in GMT-GL, every ordered pair of nodes is joined by at most one edge of every type, and
the preliminaries, including \Cref{def:weightless}, apply; the weightless graph of this construction is
$\bar\Gamma_{\mathrm{P}}$.

\textbf{Automorphisms of the graph without the weights.}
\begin{lemma}
\label{lem:structure-p}
A bijection $F : \mathcal{V} \to \mathcal{V}$ is an automorphism of the weightless graph
$\bar\Gamma_{\mathrm{P}}$ if and only if there are permutations $\pi_V \in S_V$, $\pi_d \in S_d$ and,
for every block $\ell \in [L]$, $\rho_\ell \in S_H$, $\pi^\ell_{\mathrm{ff}} \in S_{d_{\mathrm{ff}}}$
and, for every head $h \in [H]$, $\kappa_{h,\ell} \in S_{d_k}$ and $\lambda_{h,\ell} \in S_{d_v}$, such
that $F$ fixes every normalization node, acts on the nodes $v_i, v'_i, r^k_i, z^\ell_i$ as in
\Cref{eq:app-structure-gl}, and
\begin{equation}
\label{eq:app-structure-p}
F(y^{h,\ell}_i) = y^{\rho_\ell(h),\ell}_{\kappa_{h,\ell}(i)},
\qquad
F(x^{h,\ell}_i) = x^{\rho_\ell(h),\ell}_{\pi_d(i)},
\qquad
F(w^{h,\ell}_i) = w^{\rho_\ell(h),\ell}_{\lambda_{h,\ell}(i)} .
\end{equation}
We denote this map by $F^{\mathrm{P}}_\theta$, where $\theta = \bigl(\pi_V, \pi_d, (\rho_\ell,
(\kappa_{h,\ell}, \lambda_{h,\ell})_{h \in [H]}, \pi^\ell_{\mathrm{ff}})_{\ell \in [L]}\bigr)$.
\end{lemma}
Compared to GMT-GL, the only additional freedom is an independent permutation of the $d_k$ query--key
nodes per head, and of the $d_v$ value--output nodes per head.

\begin{proof}
We follow the proof of \Cref{lem:structure-gl}. Steps (1), (2) and (4)--(7), as well as the part of
step (3) concerning $\mathcal{R}^{2\ell-1}$, only involve the encodings and the coupling edge sets
between vocabulary and residual layers, which are unchanged. It remains to redo the per-head layers of
step (3), where the head nodes are now absent; their role is played by the weight edges inside each
head.

Fix a block $\ell$. The three kinds of per-head layers have different layer indices, so by the
encodings, $F$ maps each of the unions $\mathcal{Y} = \bigcup_{h} \mathcal{V}_{d_k}^{h,\ell}$,
$\mathcal{X} = \bigcup_{h} \mathcal{V}_{d}^{h,\ell}$ and $\mathcal{Z} = \bigcup_{h}
\mathcal{V}_{d_v}^{h,\ell}$ onto itself, and it remains to show that it does not split heads apart.
The edges from $\mathcal{Y}$ to $\mathcal{X}$ are exactly those of the fully-connected edge sets
$\mathcal{E}_{\crossarrow}(\mathcal{V}_{d_k}^{h,\ell}, \mathcal{V}_d^{h,\ell}; (\mathbf{W}^{k,h,\ell})^{\top})$,
$h \in [H]$: a node of $\mathcal{V}_{d_k}^{h,\ell}$ has an edge to every node of
$\mathcal{V}_d^{h,\ell}$, and to no node of $\mathcal{V}_d^{h',\ell}$ for $h' \neq h$. Likewise, the
edges from $\mathcal{X}$ to $\mathcal{Z}$ are exactly those of the edge sets
$\mathcal{E}_{\crossarrow}(\mathcal{V}_d^{h,\ell}, \mathcal{V}_{d_v}^{h,\ell}; \mathbf{W}^{v,h,\ell})$. Now fix a
head $h$, let $y \in \mathcal{V}_{d_k}^{h,\ell}$, and let $h'$ be the head with $F(y) \in
\mathcal{V}_{d_k}^{h',\ell}$. For every $x \in \mathcal{V}_{d}^{h,\ell}$, $(y, x)$ is an edge, so
$(F(y), F(x))$ is an edge from $\mathcal{V}_{d_k}^{h',\ell}$ to $\mathcal{X}$, i.e.\ $F(x) \in
\mathcal{V}_d^{h',\ell}$; hence $F$ maps $\mathcal{V}_{d}^{h,\ell}$ into $\mathcal{V}_{d}^{h',\ell}$.
Then, for every other $y' \in \mathcal{V}_{d_k}^{h,\ell}$ and $x \in \mathcal{V}_{d}^{h,\ell}$, the edge
$(F(y'), F(x))$ ends in $\mathcal{V}_d^{h',\ell}$, so $F(y') \in \mathcal{V}_{d_k}^{h',\ell}$; and for
every $z \in \mathcal{V}_{d_v}^{h,\ell}$, the edge $(F(x), F(z))$ starts in $\mathcal{V}_d^{h',\ell}$,
so $F(z) \in \mathcal{V}_{d_v}^{h',\ell}$. Thus $F$ maps the three layers of head $h$ into the
corresponding layers of a single head, which we denote $\rho_\ell(h) = h'$. Since $F$ is injective and
the corresponding layers of all heads have equal sizes, $\rho_\ell$ is injective, hence $\rho_\ell \in
S_H$, and $F$ maps every layer of head $h$ onto the corresponding layer of head $\rho_\ell(h)$. Let
$\kappa_{h,\ell}$ and $\lambda_{h,\ell}$ be the index permutations of $F$ on
$\mathcal{V}_{d_k}^{h,\ell}$ and $\mathcal{V}_{d_v}^{h,\ell}$; only fully-connected edge sets are
incident to these layers, so they are not constrained further, independently for every head and block.
On $\mathcal{V}_d^{h,\ell}$, the coupling edge set $\mathcal{E}_{\pararrow}(\mathcal{R}^{2\ell-2},
\mathcal{V}_d^{h,\ell})$ forces the index permutation $\pi_d$, exactly as in GMT-GL. Hence $F$ has the
structure of \Cref{eq:app-structure-p}, i.e.\ $F = F^{\mathrm{P}}_\theta$.

The converse is shown as in \Cref{lem:structure-gl}. Inside the attention, $F^{\mathrm{P}}_\theta$ maps
the four fully-connected edge sets of head $h$ onto those of head $\rho_\ell(h)$, which have the same
types by (ii), and the coupling edge set $\mathcal{E}_{\pararrow}(\mathcal{R}^{2\ell-2}, \mathcal{V}_d^{h,\ell})$
onto $\mathcal{E}_{\pararrow}(\mathcal{R}^{2\ell-2}, \mathcal{V}_d^{\rho_\ell(h),\ell})$, as both coupled layers
have index permutation $\pi_d$; all other edges are treated as in \Cref{lem:structure-gl}.
\end{proof}

\textbf{The action of an automorphism on the weights.}
Given $\theta$ as in \Cref{lem:structure-p}, we write $P_V$, $P_d$, $\hat{P}^\ell$ and $\sigma_\ell =
\rho_\ell^{-1}$ as for GMT-GL, and $K_{j,\ell} = P_{\kappa_{j,\ell}}$ and $\Lambda_{j,\ell} =
P_{\lambda_{j,\ell}}$.

\begin{lemma}
\label{lem:values-p}
Let $w, w' \in \mathcal{W}$, and let $F^{\mathrm{P}}_\theta$ be as in \Cref{lem:structure-p}. Then
$F^{\mathrm{P}}_\theta$ is an isomorphism from $\Gamma_{\mathrm{P}}(w)$ to $\Gamma_{\mathrm{P}}(w')$ if
and only if the first two lines of \Cref{eq:app-iso-gl} hold and, for every block $\ell$ and head $h$,
with $j = \sigma_\ell(h)$,
\begin{equation}
\label{eq:app-iso-p}
\begin{aligned}
(\mathbf{W}^{q,h,\ell})' &= P_d^{\top} \mathbf{W}^{q,j,\ell} K_{j,\ell}, &
\qquad
(\mathbf{W}^{k,h,\ell})' &= P_d^{\top} \mathbf{W}^{k,j,\ell} K_{j,\ell}, \\
(\mathbf{W}^{v,h,\ell})' &= P_d^{\top} \mathbf{W}^{v,j,\ell} \Lambda_{j,\ell}, &
\qquad
(\mathbf{W}^{O,\ell}_{h})' &= \Lambda_{j,\ell}^{\top} \mathbf{W}^{O,\ell}_{j} P_d .
\end{aligned}
\end{equation}
\end{lemma}
\begin{proof}
As in the proof of \Cref{lem:values-gl}, $F^{\mathrm{P}}_\theta$ is an isomorphism from
$\Gamma_{\mathrm{P}}(w)$ to $\Gamma_{\mathrm{P}}(w')$ if and only if every weight edge set of
$\Gamma_{\mathrm{P}}(w)$, carrying a matrix $\mathbf{M}$ between two layers on which
$F^{\mathrm{P}}_\theta$ has the index permutations $\alpha$ and $\beta$, is mapped onto an edge set of
$\Gamma_{\mathrm{P}}(w')$ carrying $\mathbf{M}' = P_\alpha^{\top} \mathbf{M} P_\beta$
(\Cref{eq:app-move-matrix}). Outside the attention sub-layers, the weight edge sets and the index
permutations on their layers are as in GMT-GL, giving the first two lines of \Cref{eq:app-iso-gl}. In
the attention of block $\ell$, $F^{\mathrm{P}}_\theta$ maps the four weight edge sets of head $j$ onto
those of head $h = \rho_\ell(j)$, with the following index permutations:
\begin{itemize}
  \item $\mathbf{W}^{q,j,\ell}$, between $\mathcal{R}^{2\ell-2}$ ($\pi_d$) and
  $\mathcal{V}_{d_k}^{j,\ell}$ ($\kappa_{j,\ell}$): $(\mathbf{W}^{q,h,\ell})' = P_d^{\top}
  \mathbf{W}^{q,j,\ell} K_{j,\ell}$;
  \item $(\mathbf{W}^{k,j,\ell})^{\top}$, between $\mathcal{V}_{d_k}^{j,\ell}$ ($\kappa_{j,\ell}$) and
  $\mathcal{V}_{d}^{j,\ell}$ ($\pi_d$): $(\mathbf{W}^{k,h,\ell})'^{\top} = K_{j,\ell}^{\top}
  (\mathbf{W}^{k,j,\ell})^{\top} P_d$, i.e.\ $(\mathbf{W}^{k,h,\ell})' = P_d^{\top} \mathbf{W}^{k,j,\ell}
  K_{j,\ell}$;
  \item $\mathbf{W}^{v,j,\ell}$, between $\mathcal{V}_{d}^{j,\ell}$ ($\pi_d$) and
  $\mathcal{V}_{d_v}^{j,\ell}$ ($\lambda_{j,\ell}$): $(\mathbf{W}^{v,h,\ell})' = P_d^{\top}
  \mathbf{W}^{v,j,\ell} \Lambda_{j,\ell}$;
  \item $\mathbf{W}^{O,\ell}_{j}$, between $\mathcal{V}_{d_v}^{j,\ell}$ ($\lambda_{j,\ell}$) and
  $\mathcal{R}^{2\ell-1}$ ($\pi_d$): $(\mathbf{W}^{O,\ell}_{h})' = \Lambda_{j,\ell}^{\top}
  \mathbf{W}^{O,\ell}_{j} P_d$.
\end{itemize}
Since $j = \sigma_\ell(h)$, these are the equations of \Cref{eq:app-iso-p}.
\end{proof}

\textbf{Proof of \Cref{thm:graph-isomorphism} for GMT-P.}
We now prove the two directions of the theorem for GMT-P, following the proof for GMT-GL.

\emph{First direction: transformers from the same orbit generate isomorphic graphs.}
Let $w \in \mathcal{W}$ and $s \in \tilde{\mathcal{G}}$, and write $w' = s \cdot w$; we build an
isomorphism from $\Gamma_{\mathrm{P}}(w)$ to $\Gamma_{\mathrm{P}}(w')$. We name the components of $s$ as
for GMT-GL. Since $s \in \tilde{\mathcal{G}}$, its general-linear components are permutation matrices,
$G^{j,\ell}_{qk} = P_{\kappa_{j,\ell}}$ and $G^{j,\ell}_{vo} = P_{\lambda_{j,\ell}}$ for unique
$\kappa_{j,\ell} \in S_{d_k}$ and $\lambda_{j,\ell} \in S_{d_v}$. We define
\[
\theta = \bigl(\pi_V, \pi_d, (\sigma_\ell^{-1}, (\kappa_{j,\ell}, \lambda_{j,\ell})_{j \in [H]},
\pi^\ell_{\mathrm{ff}})_{\ell \in [L]}\bigr),
\]
so that the matrices associated with $\theta$ in \Cref{lem:values-p} are the components of $s$, with
$K_{j,\ell} = G^{j,\ell}_{qk}$ and $\Lambda_{j,\ell} = G^{j,\ell}_{vo}$. By \Cref{lem:structure-p},
$F^{\mathrm{P}}_\theta$ is an automorphism of $\bar\Gamma_{\mathrm{P}}$. By rows (1), (2) and (4)--(7)
of \Cref{tab:symmetries}, $w$ and $w'$ satisfy the first two lines of \Cref{eq:app-iso-gl}, exactly as
for GMT-GL. For the attention, fix $\ell$ and $h$, let $j = \sigma_\ell(h)$, and abbreviate $K =
K_{j,\ell}$ and $\Lambda = \Lambda_{j,\ell}$. By row (3) of \Cref{tab:symmetries},
\begin{align*}
(\mathbf{W}^{q,h,\ell})' &= P_d^{\top} \mathbf{W}^{q,j,\ell} K, &
(\mathbf{W}^{k,h,\ell})' &= P_d^{\top} \mathbf{W}^{k,j,\ell} K^{-\top} = P_d^{\top} \mathbf{W}^{k,j,\ell}
K, \\
(\mathbf{W}^{v,h,\ell})' &= P_d^{\top} \mathbf{W}^{v,j,\ell} \Lambda, &
(\mathbf{W}^{O,\ell}_{h})' &= \Lambda^{-1} \mathbf{W}^{O,\ell}_{j} P_d = \Lambda^{\top}
\mathbf{W}^{O,\ell}_{j} P_d,
\end{align*}
since permutation matrices are orthogonal, $K^{-\top} = K$ and $\Lambda^{-1} = \Lambda^{\top}$; this is
\Cref{eq:app-iso-p}. Hence, by \Cref{lem:values-p}, $F^{\mathrm{P}}_\theta$ is an isomorphism from
$\Gamma_{\mathrm{P}}(w)$ to $\Gamma_{\mathrm{P}}(w')$, and $\Gamma_{\mathrm{P}}(w) \cong
\Gamma_{\mathrm{P}}(s \cdot w)$. In contrast to GMT-GL, the isomorphism now depends on the
general-linear components of $s$: their restriction to permutation matrices is realized by permuting the
nodes of $\mathcal{V}_{d_k}^{j,\ell}$ and $\mathcal{V}_{d_v}^{j,\ell}$.

\emph{Second direction: isomorphic graphs are generated by transformers from the same orbit.}
Let $w, w' \in \mathcal{W}$ with $\Gamma_{\mathrm{P}}(w) \cong \Gamma_{\mathrm{P}}(w')$; we prove that
$w' \in \tilde{\mathcal{G}} \cdot w$. Let $F$ be an isomorphism from $\Gamma_{\mathrm{P}}(w)$ to
$\Gamma_{\mathrm{P}}(w')$. Since, by \Cref{def:weightless}, $\Gamma_{\mathrm{P}}(w)$ and
$\Gamma_{\mathrm{P}}(w')$ have exactly the same weightless graph $\bar\Gamma_{\mathrm{P}}$, the argument
given for GMT-GL shows that $F$ is an automorphism of $\bar\Gamma_{\mathrm{P}}$. By
\Cref{lem:structure-p}, $F = F^{\mathrm{P}}_\theta$ for some $\theta$, and by \Cref{lem:values-p}, $w$
and $w'$ satisfy the first two lines of \Cref{eq:app-iso-gl} and \Cref{eq:app-iso-p}. It remains to
build from $F^{\mathrm{P}}_\theta$ a group element $s$ with $w' = s \cdot w$. Let $s$ be the element
with components $P_V$, $P_d$ and, for every block, $\sigma_\ell = \rho_\ell^{-1}$, $G^{j,\ell}_{qk} =
K_{j,\ell}$, $G^{j,\ell}_{vo} = \Lambda_{j,\ell}$ and $\hat{P}^\ell$; its general-linear components are
permutation matrices, so $s \in \tilde{\mathcal{G}}$. The first two lines of \Cref{eq:app-iso-gl} are
rows (1), (2) and (4)--(7) of \Cref{tab:symmetries}, and since $K_{j,\ell} = K_{j,\ell}^{-\top}$ and
$\Lambda_{j,\ell}^{\top} = \Lambda_{j,\ell}^{-1}$, \Cref{eq:app-iso-p} is row (3). Hence $w' = s \cdot w
\in \tilde{\mathcal{G}} \cdot w$. In contrast to GMT-GL, every attention matrix sits on its own edge
set, so the general-linear components are read off the node permutations directly, and no rank
assumption is needed.

The two directions above prove that $\Gamma_{\mathrm{P}}(w) \cong \Gamma_{\mathrm{P}}(w') \iff w' \in
\tilde{\mathcal{G}} \cdot w$, for all $w, w' \in \mathcal{W}$. Together with the GMT-GL part, this
completes the proof of \Cref{thm:graph-isomorphism}.
\hfill$\blacksquare$

\section{Experimental details}
\label{app:experimental-details}

This section collects the details of our experiments (\Cref{sec:experiments}) that are not covered in the main text: the two transformer-population datasets, the architectures of all compared metanetworks, and the training and model-selection protocol.

\subsection{Datasets}
\label{app:experimental-details-datasets}

Both datasets are populations of small character-level,
decoder-only transformer language models.
All models are of the architecture of \Cref{sec:method-symmetries}.
\Cref{tab:datasets-summary} summarizes the two populations;
we describe each in turn.

\begin{table}[H]
\caption{Summary of the two transformer-population datasets.}
\label{tab:datasets-summary}
\begin{center}
\small
\begin{tabularx}{\linewidth}{@{}l>{\raggedright\arraybackslash}X>{\raggedright\arraybackslash}X@{}}
\toprule
 & \bf Corpus-mixture dataset & \bf Fine-tuning dataset \\
\midrule
Number of models           & $3{,}000$ & $4{,}000$ \\
Parameters per model       & $91$K--$150$K (mean $121$K) & $3.8$M--$5.1$M (mean $4.4$M) \\
Vocabulary (characters)    & $64$ & $65$ \\
Embedding dim.\ $d$        & $64$ & $320$ \\
Blocks $L$                 & $\{3, 5\}$ & $6$ \\
Heads $H$                  & $\{1, 2, 4\}$ & $\{1, 2, 4\}$ \\
FFN width $d_{\mathrm{ff}}$ & $64$ & $\{320, 640\}$ \\
Dropout                    & $0.05$ & $\{0.05, 0.1, 0.2\}$ \\
Training data              & Shakespeare, TinyStories, Wikipedia & TinyShakespeare, then Wikipedia (LoRA) \\
Label(s)                   & corpus proportions & fine-tuned or not; per-weight matrix ``moved'' mask \\
\bottomrule
\end{tabularx}
\end{center}
\end{table}

\textbf{Corpus-mixture dataset.}
This dataset contains $3{,}000$ transformers trained from scratch on mixtures
of three text corpora: Shakespeare (the TinyShakespeare corpus; \citealp{karpathy2015charrnn}),
TinyStories \citep{eldan2023tinystories}, and Wikipedia (WikiText-103; \citealp{merity2017pointer}),
Before training, we ensure that all corpora have the same size by the entire dataset of the smallest of the 3 (Shakespeare), and match the number of character in the other 2 dataset (wiki + stories).
All three corpora are tokenized with a single, shared character-level vocabulary, consisting of the characters that occur in all three corpora.
Each corpus is first divided into disjoint train, validation, and test segments ($85\%/7\%/8\%$ of the characters),
and models only ever see the train segments of the corpora in their mixture.

\emph{Mixtures.} For each model we draw a vector of corpus proportions uniformly over the simplex,
and assemble the model's training text by sampling from the three corpora according to distribution vector.
Hence models differ in \emph{what} they were trained on but not in \emph{how much} data they see.
The resulting proportions are close to balanced on average ($0.33$ per corpus, ranging from below $10^{-3}$ to above $0.97$),
and the dominant corpus is nearly uniformly distributed over the population.

\emph{Architectures.} The population is architecturally diverse.
The embedding dimension and the feed-forward width are fixed,
each model independently draws its number of blocks $L \in \{3, 5\}$,
its number of attention heads $H \in \{1, 2, 4\}$
(so $d_k = d_v = d/H \in \{64, 32, 16\}$).
Models contain between $91{,}456$ and $149{,}568$ parameters ($\approx 121$K on average).

\emph{Training.} Every model is trained with next-character prediction (cross-entropy) using Adam with learning rate $10^{-3}$,
weight decay $10^{-5}$, batch size $128$, and a cosine learning-rate schedule,
for at most $30$ epochs with early stopping on the validation loss (patience $12$);
the random seed of each model is distinct (so models of identical architecture are initialized differently).

\textbf{Fine-tuning dataset.}
This dataset contains $4{,}000$ transformers, built from a population of $4{,}000$
base models that are first trained on the TinyShakespeare corpus.
Exactly half of the population ($2{,}000$ models, chosen uniformly at random) is then
LoRA fine-tuned on Wikipedia; the other half is left as-is, so each base model appears exactly once, either fine-tuned or not.

\emph{Base models.} All base models have $L = 6$ blocks and $d = 320$,
and each independently draws $H \in \{1, 2, 4\}$, $d_{\mathrm{ff}} \in \{320, 640\}$,
context length in $\{128, 256\}$, and dropout in $\{0.05, 0.1, 0.2\}$,
along with its own optimization hyperparameters: learning rate in $\{10^{-3}, 10^{-4}\}$,
weight decay in $\{0, 10^{-5}\}$, batch size in $\{64, 128, 256\}$, and a training length in $\{5, 10, 15, 20, 30\}$ epochs.
The models have between $3.78$M and $5.05$M parameters.
Training uses Adam with a cosine schedule and early stopping on validation loss (patience $12$).
Distinct random seeds are used for initialization and training, so models are initialized and trained with different random states.

\emph{LoRA fine-tuning.}
Fine-tuned models are adapted with LoRA \citep{hu2022lora} of rank $r = 4$ and scaling
$\alpha = 16$ (the update is $\tfrac{\alpha}{r}\mathbf{B}\mathbf{A}$).
The adaptable matrices are the $4$ matrices of each block
(fused query--key--value projection, attention output projection, and the two feed-forward matrices),
the embeddings, and unembedding are never adapted.
Each candidate matrix is independently selected for adaptation with probability $0.5$ (with at least one selected matrix).
Training uses AdamW with learning rate $5 \cdot 10^{-4}$, no weight decay, batch size $64$,
and a cosine schedule, for a number of epochs drawn per model from $\{3, 5, 7, 9, 11\}$
with early stopping on validation loss (patience $5$).
After fine-tuning, the low-rank updates are \emph{merged} into the base matrices,
so a fine-tuned model has exactly the same parameterization and architecture as a base one,
and no trace of the adapters remains in its weights beyond the change they induced.

\emph{Labels.} The model-level label is whether the model was fine-tuned ($2{,}000$ positives and $2{,}000$ negatives).
The weight-level label of the third experiment marks, for every weight of a fine-tuned model,
whether its value differs from the value in the base model it was derived from.

\subsection{Architectures}
\label{app:experimental-details-architectures}

The details of the GNN used by all graph constructions (GMT-P, GMT-GL, and the construction of \citet{lim2024graph,kofinas2024graph}),
together with  \citet{tran2025transformer} (NFN-T) and three flat-weight baselines (DeepSets, MLP, and MLP+Aug),
are summarized in \Cref{tab:architectures} and described below.

\begin{table}[H]
\caption{Metanetwork architectures. $N$ denotes the number of parameters of the input transformer (padded to the largest model in the population for the flat baselines). All baselines are trained with the same optimizer and protocol as the graph metanetworks (\Cref{app:experimental-details-training}).}
\label{tab:architectures}
\begin{center}
\small
\begin{tabular}{@{}llp{7.8cm}@{}}
\toprule
\bf Method & \bf Input & \bf Architecture \\
\midrule
GNN (all constructions) & graph & $3$ bidirectional GINE layers, hidden size $64$, no dropout, trainable Fourier weight encoder, MLP readout \\
NFN-T \citep{tran2025transformer} & weight tensors & one equivariant layer ($1$ channel) + invariant pooling + MLP; fixed architecture only \\
DeepSets & set of $N$ scalars & weight embedding ($128$), $3$ equivariant layers, mean-pool, MLP $[128]$, dropout $0.05$ \\
MLP & vector in $\mathbb{R}^{N}$ & hidden layers $[512, 256]$, dropout $0.1$ (reduced on the fine-tuning dataset, see text) \\
MLP+Aug & vector in $\mathbb{R}^{N}$ & same as MLP, with symmetry-group augmentation of the training weights \\
\bottomrule
\end{tabular}
\end{center}
\end{table}

\textbf{Graph metanetworks (GNN).}
All graph constructions (GMT-P, GMT-GL, and the baseline construction of \citet{lim2024graph,kofinas2024graph},
each with and without the embedding and unembedding layers)
are processed by the \emph{same} GNN within each experiment,
so that the only difference between two graph-based methods is the graph they are given.
The network follows the GINE-based design of \Cref{sec:method-architecture} and consists
of an input encoder, $3$ message-passing layers, and a readout, all with hidden size $H_{\mathrm{dim}} = 64$;
depending on the construction and the task, it has $240$K--$350$K parameters.

\emph{Input encoder.} A graph has node features $\mathbf{x}_v$ and edge features
$\mathbf{e}_{uv}$.
Each raw feature column is embedded to $\mathbb{R}^{64}$ according to its nature:
categorical types by a learned embedding table;
ordered integer indices by a linear map followed by $\sin(\cdot)$;
and the continuous parameter value $w$ by a Fourier feature encoder followed by a linear map.
The embedded parts are concatenated, layer-normalized, and linearly projected to $\mathbb{R}^{64}$,
yielding initial node states $\mathbf{h}_v^{0} \in \mathbb{R}^{64}$
and edge states $\mathbf{h}_{uv}^{0} \in \mathbb{R}^{64}$.

\emph{GINE layer.} Since transformer graphs are directed (from inputs of a parameter matrix to its outputs), each layer propagates messages both along and against the edge directions. Writing $\mathcal{N}^{\rightarrow}(v)$ and $\mathcal{N}^{\leftarrow}(v)$ for the in- and out-neighbors of $v$, layer $t$ computes
\begin{align*}
\mathbf{m}_v^{\rightarrow} &= \mathrm{MLP}^{\rightarrow}_t\!\Big( \mathbf{h}_v^{t} + \mathrm{mean}_{u \in \mathcal{N}^{\rightarrow}(v)} \mathrm{ReLU}\big(\mathbf{h}_u^{t} + \mathbf{h}_{uv}^{t}\big) \Big), \\
\mathbf{m}_v^{\leftarrow} &= \mathrm{MLP}^{\leftarrow}_t\!\Big( \mathbf{h}_v^{t} + \mathrm{mean}_{u \in \mathcal{N}^{\leftarrow}(v)} \mathrm{ReLU}\big(\mathbf{h}_u^{t} + \mathbf{h}_{vu}^{t}\big) \Big), \\
\mathbf{h}_v^{t+1} &= \mathbf{h}_v^{t} + \mathrm{ReLU}\Big(\mathrm{LN}\big(\mathbf{W}_t^{\mathrm{node}}\,[\mathbf{h}_v^{t} \,\|\, \mathbf{m}_v^{\rightarrow} \,\|\, \mathbf{m}_v^{\leftarrow}]\big)\Big), \\
\mathbf{h}_{uv}^{t+1} &= \mathbf{h}_{uv}^{t} + \mathrm{ReLU}\Big(\mathrm{LN}\big(\mathbf{W}_t^{\mathrm{edge}}\,[\mathbf{h}_u^{t} \,\|\, \mathbf{h}_v^{t} \,\|\, \mathbf{h}_{uv}^{t}]\big)\Big),
\end{align*}
Each $\mathrm{MLP}_t$ consists of two $\mathrm{Linear}\!\to\!\mathrm{LayerNorm}\!\to\!\mathrm{ReLU}$
blocks of width $64$, $\mathrm{LN}$ is layer normalization,
and $[\cdot\|\cdot]$ denotes concatenation.

\emph{Readout.} For the two model-level tasks (corpus-mixture prediction and fine-tuning detection),
the readout pools the final node states with mean and max,
and the final edge states with mean, max, and standard deviation, and concatenates the five pooled vectors.
The result is passed to a prediction MLP $\mathbb{R}^{5\cdot 64} \to \mathbb{R}^{256} \to \mathbb{R}^{128} \to \mathbb{R}^{c}$
(with LayerNorm and ReLU), where $c = 3$ for the corpus-mixture task, and $c = 2$ for fine-tuning detection.
For the weight-level task, each edge is scored independently by an MLP
$\mathbb{R}^{3\cdot 64} \to \mathbb{R}^{256} \to \mathbb{R}^{128} \to \mathbb{R}^{2}$
applied to the concatenation $[\mathbf{h}_{uv} \,\|\, \mathbf{h}_u \,\|\, \mathbf{h}_v]$
of the final state of the edge and of its two endpoints, yielding per-edge logits.
We use gradient checkpointing across the
message-passing layers to fit the multi-million-edge graphs in memory for the finetune related tasks.

\textbf{NFN-T \citep{tran2025transformer}.}
We use the architecture proposed by \citet{tran2025transformer},
which is equivariant to the attention symmetries directly in weight space, with their official implementation.
The network consists of three branches, all trained end-to-end: an equivariant branch over the attention and
feed-forward weights of the blocks, followed by an invariant pooling layer;
an MLP branch over the token embedding together with the LayerNorm parameters;
and a classifier branch over the unembedding.
We use an equivariant layer with $1$ channel,
a classifier branch with channels $[10, 10]$, and an output width of $16$ for each of the three branches;
the branch outputs are concatenated and passed to a final MLP head.

\textbf{DeepSets.}
The DeepSets baseline \citep{zaheer2017deep} reads the flattened parameter vector of a model
as an unordered set of $N$ scalars, discarding all structure.
Each scalar is embedded to $\mathbb{R}^{128}$, and processed by $3$ permutation-equivariant layers
$\mathbf{x}_i \mapsto \mathrm{ReLU}\big(\mathbf{W}\mathbf{x}_i - \mathbf{W}'\,\mathrm{mean}_j(\mathbf{x}_j)\big)$ of width $128$,
a mean-pooling over the set, and an MLP head with one hidden layer of width $128$, ReLU, and dropout $0.05$.
For the weight-level task, the pooling and head are replaced by a shared per-element head that outputs class logits
for every weight.

\textbf{MLP.}
The MLP baseline operates on the flattened parameter vector of each model,
padded with zeros to the size of the largest model in the population.
For the corpus mixture task, each scalar is first encoded to $8$ features,
and the resulting $N \times 8$ features are flattened and processed by an MLP with hidden layers
$[512, 256]$, ReLU activations, and dropout $0.1$.
For the finetune tasks we need a smaller model since the inup is much larger, making the previous selection unsuitable in memory.
Therefore each scalar is encoded to a single feature, and use hidden layers $[128, 128]$.

\textbf{MLP+Aug.}
MLP+Aug has the same architecture as the MLP, but during training the input weights are transformed by a random element of the symmetry group $\mathcal{G}$ of \Cref{eq:full-group}.
In every epoch and for every training model we independently sample a group element and apply it on the weights as in \Cref{tab:symmetries}.
Augmentation is applied during training to the training set only.

\subsection{Training schema}
\label{app:experimental-details-training}

\textbf{Common protocol.}
All methods and all three experiments share the same optimization protocol.
Models are trained with Adam and a cosine learning-rate schedule for at most $400$
epochs, and early stopping with patience of $25$ epochs:
training stops when the monitored validation metric has not improved for $25$ consecutive epochs,
and the checkpoint with the best validation value is retained.
One exception for the number of epochs is the GMT-GL variant experiment over the fine-tune detection task, where the generated graphs are particularlly very large, and so for this experiment and variant we ran for up to $130$ epochs.
The monitored metric is the validation loss (the KL divergence) for the corpus-mixture experiment,
and the validation ROC-AUC for both fine-tuning experiments.
For the corpus mixture experiment, we use a batch size of $32$ models, and for the fine-tuning experiments we use a batch size of $8$ models
(with gradient accumulation), and $16$ models for the flat baselines.

\textbf{Hyperparameter search and model selection.}
For every method and experiment, we perform a grid search over the learning rate in
$\{10^{-3}, 10^{-4}\}$ and the weight decay in $\{10^{-3}, 10^{-4}\}$; all methods share the same grid.
For each grid point, we train a model, and select the configuration(by its mean over seeds, where repeated)
with the best value of the monitored validation metric
(lowest validation KL for the corpus-mixture experiment, highest validation ROC-AUC for the fine-tuning experiments).
The numbers we report in \Cref{sec:experiments} are the test-set metrics of this selected configuration which was never used for any selection.
For the corpus-mixture experiment, each grid point is repeated with $3$ random seeds (which control the initialization and the order of training batches, with the data split fixed),
and we report mean $\pm$ standard deviation over them.

Experiments were conducted on a compute node with 8 NVIDIA L40 GPUs, each equipped with approximately 48 GB of memory

\section{Graph Construction}

\subsection{GMT-P Formal graph construction}
\label{app:graph-construction-gmt-p}

As previously mentioned, the only part where the construction GMT-P differ from GMT-GL is the MHA construction, therefore we now define exactly which edges represent the MHA weights of the model, how are they are connected, and on which edge sets corresponds to which weight matrix.

Since GMT-P restricts the symmetry to permutation matrices alone, we
no longer need to fuse each pair of projections into a single product matrix; instead, every
projection matrix corresponds directly to its own edge set. As before, we start from the node-layer
$\mathcal{V}_d$ that concludes the embedding construction (with the pre-attention normalization pair
attached exactly as in GMT-GL). For every head $h = 1, \dots, H$, we introduce
a node-layer $\mathcal{V}_{d_k}^h = \{y_i^h\}_{i=1}^{d_k}$ of $d_k$ nodes, joined to
$\mathcal{V}_d$ by the fully-connected edge set $\mathcal{E}_{\crossarrow}(\mathcal{V}_d, \mathcal{V}_{d_k}^h;
\mathbf{W}^{q,h})$, and a further node-layer $\mathcal{V}_d^h = \{x_i^h\}_{i=1}^d$ of $d$
nodes, joined to $\mathcal{V}_{d_k}^h$ by the fully-connected edge set
$\mathcal{E}_{\crossarrow}(\mathcal{V}_{d_k}^h, \mathcal{V}_d^h; (\mathbf{W}^{k,h})^\top)$. As with the GMT-GL, and for the same reasons, we tie the permutation of $\mathcal{V}_d^h$ to that of $\mathcal{V}_d$
with the coupling edge set $\mathcal{E}_{\pararrow}(\mathcal{V}_d, \mathcal{V}_d^h)$.
Each per-head layer $\mathcal{V}_d^h$ is then joined to a node-layer $\mathcal{V}_{d_v}^h =
\{w_i^h\}_{i=1}^{d_v}$ of $d_v$ nodes by the fully-connected edge set
$\mathcal{E}_{\crossarrow}(\mathcal{V}_d^h, \mathcal{V}_{d_v}^h; \mathbf{W}^{v,h})$, which in turn is joined to a
single shared node-layer $\mathcal{V}_d' = \{x_i'\}_{i=1}^d$ of $d$ nodes by the
fully-connected edge set $\mathcal{E}_{\crossarrow}(\mathcal{V}_{d_v}^h, \mathcal{V}_d'; \mathbf{W}^{O,h})$,
writing $\mathbf{W}^{O,h}$ for the $h$-th row-block of $\mathbf{W}^{O}$, and merging the
heads back into the residual stream. Finally, we add the
coupling edge set $\mathcal{E}_{\pararrow}(\mathcal{V}_d, \mathcal{V}_d')$ to account for the residual
connection wrapped around the MHA sub-layer. The remainder of the construction 
is identical to that of GMT-GL.

\subsection{Complete 1 block LM graph constructions}
\label{app:graph-construction-figures}

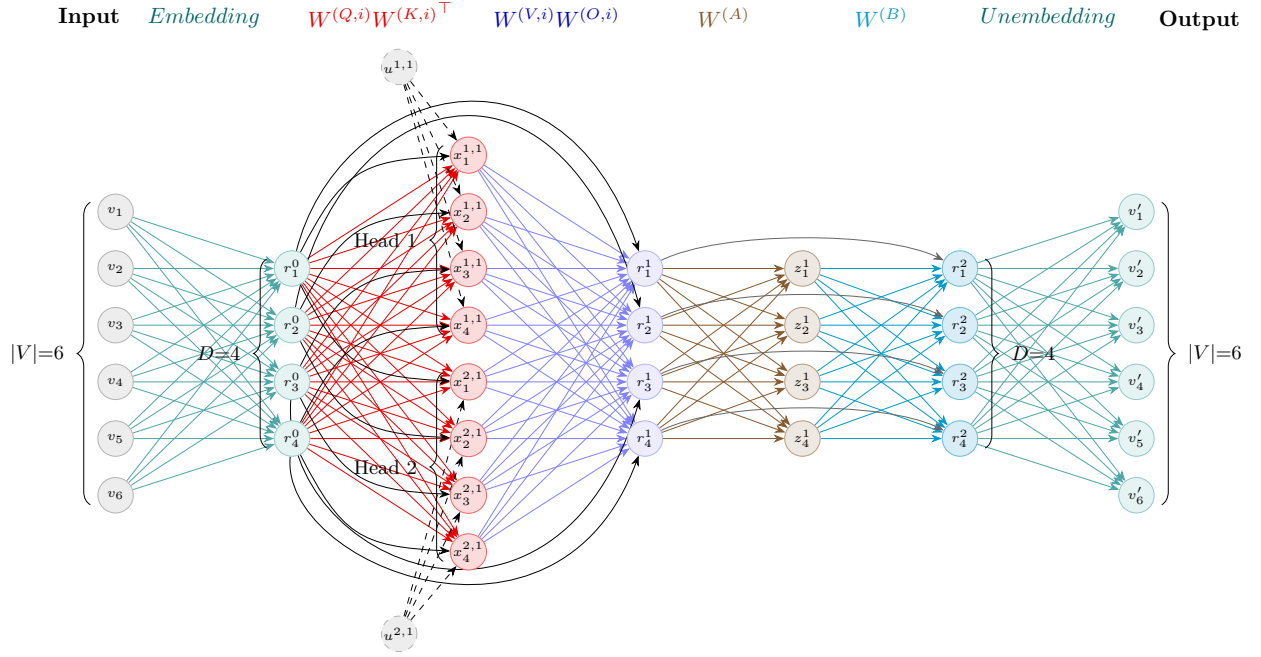
\begin{figure}[H]
\centering
\colorlet{gmtV}{black!50}         
\colorlet{gmtRa}{teal!70}         
\colorlet{gmtU}{black!50}         
\colorlet{gmtX}{red!90!black}     
\colorlet{gmtRb}{blue!50}         
\colorlet{gmtZ}{brown!70!black}   
\colorlet{gmtRc}{cyan!80!black}   
\colorlet{gmtVp}{teal!70}         
\resizebox{\linewidth}{!}{%
\begin{tikzpicture}[
    node distance=3.2cm and 3.5cm,
    neuron/.style={circle, draw, minimum size=16pt, inner sep=0pt, fill=white, font=\tiny},
    layer/.style={neuron, fill=#1!14, draw=#1!60},
    >=Stealth,
    every edge/.style={draw, ->},
    brace/.style={decorate, decoration={brace, amplitude=6pt, mirror}},
    brace above/.style={decorate, decoration={brace, amplitude=6pt, mirror}},
    label font/.style={font=\footnotesize},
]


\foreach \i in {1,...,6} {
    \node[layer=gmtV] (L0-\i) at (0, -\i*0.9 - 1.4) {$v_{\i}$};
}

\foreach \i in {1,...,4} {
    \node[layer=gmtRa] (L1-\i) at (2.8, -\i*0.9 - 2.3) {$r^{0}_{\i}$};
}

\foreach \i in {1,...,4} {
    \node[layer=gmtX] (L3H1-\i) at (5.6, -\i*0.9 - 0.5) {$x^{1,1}_{\i}$};
}
\foreach \i in {1,...,4} {
    \node[layer=gmtX] (L3H2-\i) at (5.6, -\i*0.9 - 4.1) {$x^{2,1}_{\i}$};
}

\foreach \i in {1,...,4} {
    \node[layer=gmtRb] (L5-\i) at (8.4, -\i*0.9 - 2.3) {$r^{1}_{\i}$};
}

\foreach \i in {1,...,4} {
    \node[layer=gmtZ] (L6-\i) at (10.9, -\i*0.9 - 2.3) {$z^{1}_{\i}$};
}
\foreach \i in {1,...,4} {
    \node[layer=gmtRc] (L7-\i) at (13.4, -\i*0.9 - 2.3) {$r^{2}_{\i}$};
}

\foreach \i in {1,...,6} {
    \node[layer=gmtVp] (LU-\i) at (16.2, -\i*0.9 - 1.4) {$v'_{\i}$};
}

\begin{pgfonlayer}{background}
\foreach \i in {1,...,6} {
    \foreach \j in {1,...,4} {
        \draw[->, teal!70] (L0-\i) -- (L1-\j);
    }
}
\foreach \i in {1,...,4} {
    \foreach \j in {1,...,4} {
        \draw[->, red!90!black] (L1-\i) -- (L3H1-\j);
        \draw[->, red!90!black] (L1-\i) -- (L3H2-\j);
    }
}
\foreach \i in {1,...,4} {
    \foreach \j in {1,...,4} {
        \draw[->, blue!50] (L3H1-\i) -- (L5-\j);
        \draw[->, blue!50] (L3H2-\i) -- (L5-\j);
    }
}
\foreach \i in {1,...,4} {
    \draw[->, thin, solid, black] (L1-\i) to[bend left=30, looseness=1.3] (L3H1-\i);
    \draw[->, thin, solid, black] (L1-\i) to[bend right=30, looseness=1.3] (L3H2-\i);
}
\foreach \i in {1,...,2} {
    \draw[->, thin, solid, black] (L1-\i) .. controls (3.5, -0.4) and (7.0, 0.8) .. (L5-\i);
}
\foreach \i in {3,...,4} {
    \draw[->, thin, solid, black] (L1-\i) .. controls (2.6, -7.9) and (7.0, -9.7) .. (L5-\i);
}
\end{pgfonlayer}
\node[layer=gmtU, dashed] (UH1) at (4.5, 0.0) {$u^{1,1}$};
\node[layer=gmtU, dashed] (UH2) at (4.5, -9.0) {$u^{2,1}$};
\begin{pgfonlayer}{background}
\foreach \i in {1,...,4} {
    \draw[->, thin, dashed, black] (UH1) -- (L3H1-\i);
    \draw[->, thin, dashed, black] (UH2) -- (L3H2-\i);
}
\foreach \i in {1,...,4} {
    \foreach \j in {1,...,4} {
        \draw[->, brown!70!black] (L5-\i) -- (L6-\j);
    }
}
\foreach \i in {1,...,4} {
    \foreach \j in {1,...,4} {
        \draw[->, cyan!80!black] (L6-\i) -- (L7-\j);
    }
}
\foreach \i in {1,...,2} {
    \draw[->, thin, solid, black!60] (L5-\i) .. controls +(1.2,0.6) and +(-1.2,0.6) .. (L7-\i);
}
\foreach \i in {3,...,4} {
    \draw[->, thin, solid, black!60] (L5-\i) .. controls +(1.2,0.6) and +(-1.2,0.6) .. (L7-\i);
}
\foreach \i in {1,...,4} {
    \foreach \j in {1,...,6} {
        \draw[->, teal!70] (L7-\i) -- (LU-\j);
    }
}
\end{pgfonlayer}

\draw[brace above] (-0.4, -2.3 + 0.15) -- (-0.4, -6.8 - 0.15)
    node[midway, left=8pt, font=\footnotesize] {$|V|{=}6$};
\draw[brace above] (2.4, -3.2 + 0.15) -- (2.4, -5.9 - 0.15)
    node[midway, left=8pt, font=\footnotesize] {$D{=}4$};
\draw[brace above] (5.2, -1.4 + 0.15) -- (5.2, -4.1 - 0.15)
    node[midway, left=8pt, font=\footnotesize] {Head 1};
\draw[brace above] (5.2, -5.0 + 0.15) -- (5.2, -7.7 - 0.15)
    node[midway, left=8pt, font=\footnotesize] {Head 2};
\draw[brace] (13.8, -5.9 - 0.15) -- (13.8, -3.2 + 0.15)
    node[midway, right=8pt, font=\footnotesize] {$D{=}4$};
\draw[brace] (16.6, -6.8 - 0.15) -- (16.6, -2.3 + 0.15)
    node[midway, right=8pt, font=\footnotesize] {$|V|{=}6$};

\node[above, font=\small\bfseries] at (-0.42, 0.5) {Input};
\node[above, font=\small\bfseries, text=teal!80!black] at (1.4, 0.5) {$Embedding$};
\node[above, font=\small\bfseries, text=red!90!black] at (4.2, 0.5) {$W^{(Q,i)} {W^{(K,i)}}^\top$};
\node[above, font=\small\bfseries, text=blue!80!black] at (7.0, 0.5) {$W^{(V,i)} W^{(O,i)}$};
\node[above, font=\small\bfseries, text=brown!70!black] at (9.65, 0.5) {$W^{(A)}$};
\node[above, font=\small\bfseries, text=cyan!80!black] at (12.15, 0.5) {$W^{(B)}$};
\node[above, font=\small\bfseries, text=teal!80!black] at (14.8, 0.5) {$Unembedding$};
\node[above, font=\small\bfseries] at (17.2, 0.45) {Output};

\end{tikzpicture}%
}
\caption{Full construction of a single-block transformer, containing a 2-head attention mechanism. We omit the coupling between the first and last layers, as well as the normalization nodes and edges, to keep the plot intuitive and readable.}
\label{fig:GL-symm-1-block}
\end{figure}

\begin{figure}[H]
\centering
\colorlet{gmtV}{black!50}         
\colorlet{gmtRa}{teal!70}         
\colorlet{gmtY}{gray!60}          
\colorlet{gmtX}{orange!70}        
\colorlet{gmtW}{blue!50}          
\colorlet{gmtRb}{violet!70}       
\colorlet{gmtZ}{brown!70!black}   
\colorlet{gmtRc}{cyan!80!black}   
\colorlet{gmtVp}{teal!70}         
\resizebox{\linewidth}{!}{%
\begin{tikzpicture}[
    node distance=3.2cm and 3.5cm,
    neuron/.style={circle, draw, minimum size=16pt, inner sep=0pt, fill=white, font=\tiny},
    layer/.style={neuron, fill=#1!14, draw=#1!60},
    >=Stealth,
    every edge/.style={draw, ->},
    brace/.style={decorate, decoration={brace, amplitude=6pt, mirror}},
    brace above/.style={decorate, decoration={brace, amplitude=6pt, mirror}},
    label font/.style={font=\footnotesize},
]


\foreach \i in {1,...,6} {
    \node[layer=gmtV] (L0-\i) at (0, -\i*0.9 - 1.4) {$v_{\i}$};
}

\foreach \i in {1,...,4} {
    \node[layer=gmtRa] (L1-\i) at (2.1, -\i*0.9 - 2.3) {$r^{0}_{\i}$};
}

\foreach \i in {1,...,2} {
    \node[layer=gmtY] (L2H1-\i) at (3.9, -\i*0.9 - 1.4) {$y^{1,1}_{\i}$};
}
\foreach \i in {1,...,2} {
    \node[layer=gmtY] (L2H2-\i) at (3.9, -\i*0.9 - 5.0) {$y^{2,1}_{\i}$};
}

\foreach \i in {1,...,4} {
    \node[layer=gmtX] (L3H1-\i) at (5.7, -\i*0.9 - 0.5) {$x^{1,1}_{\i}$};
}
\foreach \i in {1,...,4} {
    \node[layer=gmtX] (L3H2-\i) at (5.7, -\i*0.9 - 4.1) {$x^{2,1}_{\i}$};
}

\foreach \i in {1,...,2} {
    \node[layer=gmtW] (L4H1-\i) at (7.5, -\i*0.9 - 1.4) {$w^{1,1}_{\i}$};
}
\foreach \i in {1,...,2} {
    \node[layer=gmtW] (L4H2-\i) at (7.5, -\i*0.9 - 5.0) {$w^{2,1}_{\i}$};
}

\foreach \i in {1,...,4} {
    \node[layer=gmtRb] (L5-\i) at (9.3, -\i*0.9 - 2.3) {$r^{1}_{\i}$};
}

\foreach \i in {1,...,4} {
    \node[layer=gmtZ] (L6-\i) at (11.4, -\i*0.9 - 2.3) {$z^{1}_{\i}$};
}
\foreach \i in {1,...,4} {
    \node[layer=gmtRc] (L7-\i) at (13.5, -\i*0.9 - 2.3) {$r^{2}_{\i}$};
}

\foreach \i in {1,...,6} {
    \node[layer=gmtVp] (LU-\i) at (15.6, -\i*0.9 - 1.4) {$v'_{\i}$};
}

\begin{pgfonlayer}{background}
\foreach \i in {1,...,6} {
    \foreach \j in {1,...,4} {
        \draw[->, teal!70] (L0-\i) -- (L1-\j);
    }
}
\foreach \i in {1,...,4} {
    \foreach \j in {1,...,2} {
        \draw[->, gray!60] (L1-\i) -- (L2H1-\j);
        \draw[->, gray!60] (L1-\i) -- (L2H2-\j);
    }
}
\foreach \i in {1,...,2} {
    \foreach \j in {1,...,4} {
        \draw[->, orange!70] (L2H1-\i) -- (L3H1-\j);
        \draw[->, orange!70] (L2H2-\i) -- (L3H2-\j);
    }
}
\foreach \i in {1,...,4} {
    \foreach \j in {1,...,2} {
        \draw[->, blue!50] (L3H1-\i) -- (L4H1-\j);
        \draw[->, blue!50] (L3H2-\i) -- (L4H2-\j);
    }
}
\foreach \i in {1,...,2} {
    \foreach \j in {1,...,4} {
        \draw[->, violet!70] (L4H1-\i) -- (L5-\j);
        \draw[->, violet!70] (L4H2-\i) -- (L5-\j);
    }
}
\foreach \i in {1,...,4} {
    \draw[->, thin, solid, black!65] (L1-\i) .. controls +(1.2,0.9) and +(-1.2,0.3) .. (L3H1-\i);
    \draw[->, thin, solid, black!65] (L1-\i) .. controls +(1.2,-0.9) and +(-1.2,-0.3) .. (L3H2-\i);
}
\foreach \i in {1,...,2} {
    \draw[->, thin, solid, black!50] (L1-\i) .. controls (3.0, -0.4) and (7.5, 0.8) .. (L5-\i);
}
\foreach \i in {3,...,4} {
    \draw[->, thin, solid, black!50] (L1-\i) .. controls (1.83, -7.9) and (7.5, -9.7) .. (L5-\i);
}
\foreach \i in {1,...,4} {
    \foreach \j in {1,...,4} {
        \draw[->, brown!70!black] (L5-\i) -- (L6-\j);
    }
}
\foreach \i in {1,...,4} {
    \foreach \j in {1,...,4} {
        \draw[->, cyan!80!black] (L6-\i) -- (L7-\j);
    }
}
\foreach \i in {1,...,2} {
    \draw[->, thin, solid, black!60] (L5-\i) .. controls +(1.2,0.6) and +(-1.2,0.6) .. (L7-\i);
}
\foreach \i in {3,...,4} {
    \draw[->, thin, solid, black!60] (L5-\i) .. controls +(1.2,0.6) and +(-1.2,0.6) .. (L7-\i);
}
\foreach \i in {1,...,4} {
    \foreach \j in {1,...,6} {
        \draw[->, teal!70] (L7-\i) -- (LU-\j);
    }
}
\end{pgfonlayer}

\draw[brace above] (-0.4, -2.3 + 0.15) -- (-0.4, -6.8 - 0.15)
    node[midway, left=8pt, font=\footnotesize] {$|V|{=}6$};
\draw[brace above] (1.7, -3.2 + 0.15) -- (1.7, -5.9 - 0.15)
    node[midway, left=8pt, font=\footnotesize] {$D{=}4$};
\draw[brace above] (3.5, -2.3 + 0.15) -- (3.5, -3.2 - 0.15)
    node[midway, left=8pt, font=\footnotesize] {Head 1};
\draw[brace above] (3.5, -5.9 + 0.15) -- (3.5, -6.8 - 0.15)
    node[midway, left=8pt, font=\footnotesize] {Head 2};
\draw[brace] (13.9, -5.9 - 0.15) -- (13.9, -3.2 + 0.15)
    node[midway, right=8pt, font=\footnotesize] {$D{=}4$};
\draw[brace] (16.0, -6.8 - 0.15) -- (16.0, -2.3 + 0.15)
    node[midway, right=8pt, font=\footnotesize] {$|V|{=}6$};

\node[above, font=\small\bfseries] at (-0.42, 0.5) {Input};
\node[above, font=\small\bfseries, text=teal!80!black] at (1.05, 0.5) {$Embedding$};
\node[above, font=\small\bfseries, text=gray!90] at (3.0, 0.5) {$W^{(Q,i)}$};
\node[above, font=\small\bfseries, text=orange!80!black] at (4.8, 0.5) {${W^{(K,i)}}^\top$};
\node[above, font=\small\bfseries, text=blue!80!black] at (6.6, 0.5) {$W^{(V,i)}$};
\node[above, font=\small\bfseries, text=violet!80!black] at (8.4, 0.5) {$W^{(O,i)}$};
\node[above, font=\small\bfseries, text=brown!70!black] at (10.35, 0.5) {$W^{(A)}$};
\node[above, font=\small\bfseries, text=cyan!80!black] at (12.45, 0.5) {$W^{(B)}$};
\node[above, font=\small\bfseries, text=teal!80!black] at (14.55, 0.5) {$Unembedding$};
\node[above, font=\small\bfseries] at (16.6, 0.45) {Output};

\end{tikzpicture}%
}
\caption{Full construction of a single-block transformer,
containing a 2-head attention mechanism. Attention construction is invariant to permutations only.
We omit the coupling between the first and last layers, as well as the normalization nodes and edges, to keep the plot intuitive and readable.}
\label{fig:P-symm-1-block}
\end{figure}
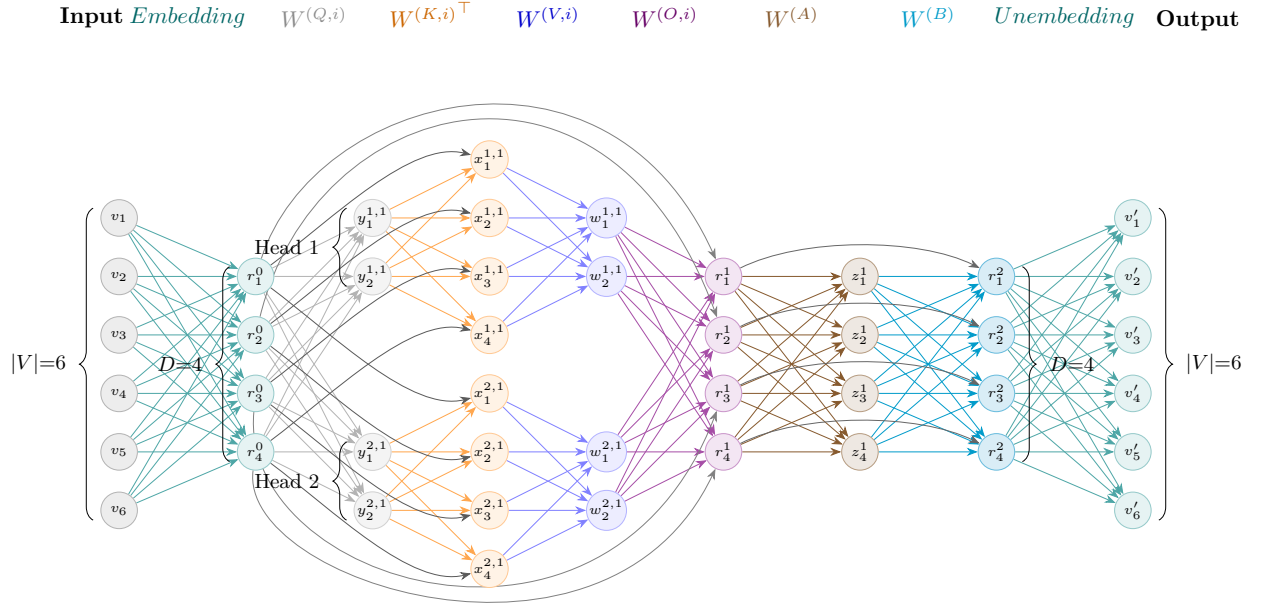

To better understand how the full graphs are constructed from a model, let us present also the full construction of a language model with 1 transformer block, for both constructions GMT-GL \Cref{fig:GL-symm-1-block} and GMT-P  \Cref{fig:P-symm-1-block}.

\end{document}